%% file: main.tex
\documentclass[sigconf]{acmart}

\usepackage{booktabs}
\usepackage{subfig}
\usepackage{bm}
\usepackage{balance}
\usepackage{tikz}
\usepackage{colortbl}
\usepackage{subfig}
\usepackage{graphicx}
\usetikzlibrary{positioning}
\usepackage{enumitem}

\usepackage{amsthm}
\usepackage{amsmath}
\usepackage{amsfonts}
\usepackage{bbm}
\usepackage{chngcntr}
\usepackage{apptools}

\newtheorem*{theorem*}{Theorem}

\newtheorem*{lemma*}{Lemma}

\newtheorem*{claim*}{Claim}

\newtheorem*{proposition*}{Proposition}
\DeclareMathOperator*{\argmax}{arg\,max}

\setcopyright{rightsretained}
\renewcommand\footnotetextcopyrightpermission[1]{}
\acmConference[]{}{}{}
\author{Sebastian Bruch, Jan Pfeifer, Mathieu Guillame-Bert}
\authornote{All authors contributed equally.}
\affiliation{%
  \institution{Google Research}
  \streetaddress{}
  \city{} 
  \state{} 
  \postcode{}
}
\email{{bruch,janpf,gbm}@google.com}

\begin{document}
\title[Learning Representations for Axis-Aligned Decision Forests]{Learning Representations for Axis-Aligned \\
Decision Forests through Input Perturbation}

\begin{abstract}
Axis-aligned decision forests have long been the leading class of machine learning algorithms for modeling tabular data. In many applications of machine learning such as learning-to-rank, decision forests deliver remarkable performance. They also possess other coveted characteristics such as interpretability. Despite their widespread use and rich history, decision forests to date fail to consume raw structured data such as text, or learn effective representations for them, a factor behind the success of deep neural networks in recent years. While there exist methods that construct smoothed decision forests to achieve representation learning, the resulting models are decision forests in name only: They are no longer axis-aligned, use stochastic decisions, or are not interpretable. Furthermore, none of the existing methods are appropriate for problems that require a Transfer Learning treatment. In this work, we present a novel but intuitive proposal to achieve representation learning for decision forests without imposing new restrictions or necessitating structural changes. Our model is simply a decision forest, possibly trained using any forest learning algorithm, atop a deep neural network. By approximating the gradients of the decision forest through input perturbation, a purely analytical procedure, the decision forest directs the neural network to learn or fine-tune representations. Our framework has the advantage that it is applicable to any arbitrary decision forest and that it allows the use of arbitrary deep neural networks for representation learning. We demonstrate the feasibility and effectiveness of our proposal through experiments on synthetic and benchmark classification datasets.
\end{abstract}




\maketitle

\input{introduction}
\input{related_work}
\input{preliminaries}
\input{proposed_method}
\input{experiments}
\input{conclusion}

\newpage
\balance
\bibliographystyle{ACM-Reference-Format}
\bibliography{main} 

\end{document}

%% file: introduction.tex
\section{Introduction} \label{sec:introduction}

Ensembles of decision trees, known as decision forests, such as Random Forests~\cite{Breiman2001Randomforests} and Gradient Boosted Decision Trees~\cite{friedman2001greedy} (GBDTs) enjoy a considerable degree of prominence among machine learning methods. They have attained their high status owing to a variety of reasons including their ability to produce \emph{interpretable} models as well as their power to model \emph{tabular} data. For certain tasks, the effectiveness of models learnt by decision forests is arguably unparalleled: in the domain of learning-to-rank, for example, this has been demonstrated time and again~\cite{BruchApproxSIGIR2019,Pasumarthi:arxiv:1910.09676,Zhuang:arxiv:2005.02553,Bruch:wsdm:2020}, with GBDTs serving as a basis for the leading ranking functions such as LambdaMART~\cite{burges2010ranknet,wu2010adapting} and $\textsc{XE}_\textsc{NDCG}$~\cite{Bruch2019XEndcg}.

Despite their success on many fronts, processing raw structured data such as text, audio, or image for perceptual tasks has long remained an elusive target for decision forests. Where decision forests prove ineffective, however, Deep Learning~\cite{Goodfellow2016DeepLearning} has filled the gap with remarkable success. A key factor driving that success and popularity is argued to be the ability of deep neural networks to learn compelling representations of raw structured data~\cite{Begio2013RepresentationLearning}---hereafter referred to as \emph{embeddings}. The mechanics of learning an embedding or adapting (or \emph{fine-tuning}) a \emph{pre-trained} embedding, is rather trivial with gradient descent optimization methods, and is made possible largely thanks to the differentiability of neural networks. Differentiability is, however, a property decision forests famously do not possess due to their discontinuous structure.

Bridging the gap between decision forests and Deep Learning bears significant importance to many applications, and has unsurprisingly inspired an array of ``hybrid'' solutions~\cite{Li2019CombiningDTandNNs,Ke2019DeepGBM,Richmond2016MappingAutoContext}. The goal is to leverage the powers of the two by placing them alongside each other, all to ultimately model heterogeneous datasets that comprise of engineered features (consumed by the decision forest) as well as raw structured data (consumed separately by the neural network). But hybrid models require much engineering and necessitate much attention to the interplay between the two disparate building blocks: Embeddings learnt by the neural network component, for example, are detached from the decision forest block. These pitfalls bring us to a research topic that also piques academic interest: Designing decision forests that are equipped to learn embeddings.

Whether and how a decision tree or a decision forest can be formulated to \emph{drive} the process of learning embeddings in spite of their inherent non-differentiability, is indeed a question that has lately become of interest to the research community~\cite{Feng2018MultiLayeredGBDTs,Zhou2017DeepForest,Kontschieder2016DNDF,Yang2018DNDT,Wan2020NBDT,Balestriero2017NDT}. Its pursuit is rightly justified by not just a desire to apply decision forests to raw structured data, but also to minimize the complexity of the underlying optimization problem and its parameter space as compared with hybrid models, or to induce interpretability in the final model. A similar set of factors motivate us to examine this same question in this work.

A related but unexplored research question is whether and how one may carry over pre-trained embeddings to a decision forest model---a scenario not directly nor trivially supported by methods noted above, methods that are designed with the express purpose of learning representations from scratch. And that inability to share and fine-tune data representations, in turn, hinders the use of decision forests or renders them ineffective for problems that require treatments such as Transfer Learning (e.g., in settings where there is a paucity of training data to learn effective embeddings from scratch). These scenarios are not only interesting from an academic standpoint, but also are common in practice. We therefore believe this research question to be crucial and investigate it in this work.

Our work on the two research questions above---how to bestow the ability to drive (a) \emph{learning} or (b) \emph{fine-tuning} of embeddings to decision forests---has led us to a solution that is intuitive, is straightforward to implement, and, perhaps more remarkably, may be applied to any decision forest.

At a high level, our proposal is a decision forest atop an embedding function---a neural network. The decision forest is randomly generated when we intend to \emph{learn} embeddings from scratch. When \emph{fine-tuning} pre-trained embeddings, however, it is a Random Forest or a GBDT trained on those initial embeddings. Regardless of how the decision forest is produced, we must facilitate back-propagation in order to direct the neural network to learn appropriate embeddings. We do so by \emph{approximating} the gradient of the decision forest with respect to its input through a process of input perturbation: Input to the decision forest is perturbed with a zero-mean noise distribution and the output of the decision forest is taken to be its expected value over the perturbed input. This process to approximate the gradients is purely analytical, and, as such, does not require any change to the structure of the decision forest.

We show how our proposed gradient approximation method has a similar effect as a smoothing of decision boundaries. We subsequently put our proposal to the test through a series of experiments with synthetically generated data as well as benchmark datasets. The results reported in this work demonstrate the effectiveness of this setup both at learning embeddings and at fine-tuning pre-trained embeddings.

Our proposal differs from previous work in three notable ways. First and foremost, the decision forest component of our solution remains a proper decision forest---with hard splits in intermediate nodes, enabling utilization of efficient tree inference algorithms~\cite{Asadi2014RuntimeOpt,Lucchese2015QuickScorer,Lettich2019ParallelTraversal}. Most existing approaches, on the other hand, define a decision forest rather liberally by including ``soft'' or smoothed variants of it in their definition or by allowing oblique decision boundaries. Second, as we noted earlier, our methodology can be used to fine-tune pre-trained embeddings, enabling its use in applications that require Transfer Learning. Lastly, in our solution, the decision forest makes no distinction between tabular features and embeddings; both types of input features may be used together to train the decision forest.

The following list summarizes our contributions:
\begin{itemize}
    \item We introduce a framework where a decision forest harvests Deep Learning components to learn embeddings;
    \item We study the behavior of our proposal through experiments with synthetic data; and,
    \item We demonstrate through extensive experiments on benchmark datasets the effectiveness of our proposed method on learning or fine-tuning embeddings.
\end{itemize}

The remainder of this paper is organized as follows: In Section~\ref{sec:related_work} we review the literature and contrast our proposal with past work. Section~\ref{sec:preliminaries} sets up the notation we adopt in this work and covers background material. Our proposed method is given in detail in Section~\ref{sec:proposed_method}. Section~\ref{sec:experiments} presents and discusses the results of our experiments on synthetic and benchmark datasets. Finally, we conclude this work in Section~\ref{sec:conclusion} and elaborate our future plans.

%% file: related_work.tex
\section{Related Work} \label{sec:related_work}

Deep Learning has in recent years transformed the landscape of machine learning with many of its applications making great strides as a result. In Natural Language Processing, Question Answering, and Learning-to-Rank tasks, for example, dramatic findings have been reported (e.g., ~\cite{Nogueira2019multistage,Han2020learningtorank,devlin2018bert}) as a direct consequence of this new ability to consume text by way of representation learning.

It is of little surprise then that researchers began to investigate ways to combine the power of deep neural networks to learn effective embeddings, with the ability of decision forests to learn effective models for various tasks~\cite{lightgbm2017nips,Chen2016XGBoost}. In its simplest form, the two coexist alongside each other where one (neural network) is delegated with learning embeddings for ``sparse'' categorical features while the other (decision forest) models the ``dense'' numerical features. This is precisely what Ke et al.~\cite{Ke2019DeepGBM} have recently proposed. In a similar manner, though with a rather different purpose, Li et al.~\cite{Li2019CombiningDTandNNs} devised a setup where a neural network is trained first, then a decision forest is trained to model the residual errors, or vice versa.

The decoupling of (differentiable) neural networks and (non-differentiable) decision forests is sensible as the training algorithms for the two are incompatible. Though, it leaves one with two components that have rather disparate objectives to optimize. For example, the learning of embeddings by the neural network is not directly guided or influenced by the decision forest.

One solution is to forgo Deep Learning and explore ways in which decision forests may, on their own, learn embeddings. This question was first raised by Zhou and Feng in~\cite{Zhou2017DeepForest} and investigated further in~\cite{Feng2018MultiLayeredGBDTs}: The idea is to use ensembles of decision forests, or a hierarchy thereof, to learn embeddings. While these methods introduce an exciting line of research and demonstrate promising results, they fail to leverage existing findings from the Deep Learning literature; embeddings cannot be transferred from neural networks, and must instead be learnt from scratch by decision forests. We, on the other hand, focus on a solution that is capable of doing both.

Another class of solutions offered in the literature involve a ``softening'' of the structure of a decision tree---a solution that could presumably generalize to decision forests. Once a decision tree is ``soft'' or differentiable, it may be optimized jointly with a neural network. Balestriero~\cite{Balestriero2017NDT} achieves this by considering oblique decision boundaries as opposed to the traditional axis-aligned hyperplanes. Kontschieder et al.~\cite{Kontschieder2016DNDF} instead use stochastic splits---a decision at every node to take the left or right branch is made according to a probability distribution. A similar approach is considered in the work of Yang et al.~\cite{Yang2018DNDT} and Su{\'a}rez and Lutsko~\cite{Surez1999GloballyOF}.

The methods in this class have the disadvantage that their output is an oblique or a soft tree---a form that deviates greatly from traditional decision forests and cannot take advantage of advanced inference algorithms~\cite{Lucchese2015QuickScorer} and other technologies tailored to the traditional tree structure. It is also not immediately clear how these methods could leverage pre-trained embeddings in a Transfer Learning scenario. Finally, the proposal in \cite{Yang2018DNDT} does not scale to large numbers of features or large forests due to its use of the Kronecker product, an expensive algorithm. Our work, in contrast, exhibits none of these shortcomings.

Another recent work on this topic is \cite{Wan2020NBDT}, though in their work Wan et al. investigate an entirely different research question: Explaining the decision boundaries of a neural network using decision trees. The authors devise an algorithm to induce a decision tree given the weights of a neural network. This approach does not, directly or indirectly, serve as a suitable solution to the research questions raised in this work.

Lastly, as we will explain in Section~\ref{sec:proposed_method}, the idea that lies at the heart of our method can be summarized as input perturbation. Perturbing the input to a function for smoothing purposes is not itself a novel contribution. SoftRank~\cite{Taylor2008SoftRank}, for example, is a learning-to-rank method where a ranking metric is made differentiable by way of perturbing ranking scores with Gaussian noise. Bruch et al. in~\cite{Bruch:wsdm:2020} sample ranking scores from a distribution formed by Gumbel noise. Berthet et al.~\cite{Berthet2020PerturbedOptimizers} make certain non-differentiable objectives differentiable by injecting noise into solvers. There exist many other examples, but, to the best of our knowledge, we are the first to utilize this technique in the context of decision forests for the purpose of learning embeddings.

%% file: preliminaries.tex
\section{Preliminaries}\label{sec:preliminaries}

This section sets up our notation and goes over a few basic concepts to set the stage for a more in-depth discussion in future sections. We begin with a brief review of decision trees.

A \emph{decision tree} is a tree structure---not necessarily balanced or complete---with a number of \emph{intermediate} nodes and \emph{leaf} nodes (or \emph{leaves} for brevity). In this work, we limit our focus to binary decision trees, a structure that is more common in practice. In a binary decision tree, every intermediate node contains a yes-or-no rule, according to which one proceeds either to the left or the right branch. This is referred to as a \emph{split} or a \emph{decision}. In this work, as is often the case elsewhere, we only consider axis-aligned decisions: A decision is made by comparing a single numerical feature with a threshold (i.e., $x_i \ge \theta$, for some feature $x_i$ and threshold $\theta$). Every leaf of a decision tree has a value which, for example, may be numerical in \emph{regression trees}, binary for binary classification trees, a probability distribution in multi-class classification trees, etc. In other words, intermediate nodes partition the space into hypercubes (regions) and leaves define the value of the function in each hypercube (region).

Let us denote a decision tree with $\mathcal{T}$ and its partition of the $m$-dimensional input space with $\mathcal{R_T}$. Concretely, $\mathcal{R_T}$ comprises of disjoint axis-aligned regions, one per leaf, whose union is $\mathbb{R}^m$. Let us also denote a leaf value with $\mathcal{L_T}(R)$, where $R \in \mathcal{R_T}$ is a single region, and an $m$-dimensional input with $\bm{x} = (x_1, x_2, \ldots, x_m) \in \mathbb{R}^m$. Then the output of a decision tree can be expressed compactly as follows:
\begin{equation}
    \mathcal{T}(\bm{x}) \triangleq \sum_{R \in \mathcal{R_T}} \mathcal{L_T}(R) \mathbbm{1}_{\bm{x} \in R},
    \label{equ:decision_tree_function}
\end{equation}
where $\mathbbm{1}_{c}$ is the indicator function taking on the value 1 if the condition $c$ is true and 0 otherwise. Note that, in decision trees with axis-aligned decisions, an input $\bm{x}$ belongs to exactly one region.

\begin{figure}
\begin{center}
\centerline{
\subfloat[Decision Forest]{\includegraphics[width=2.3in]{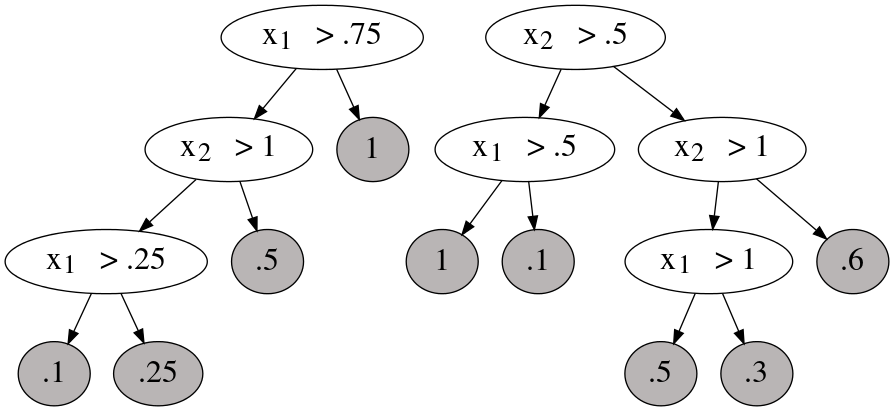}}
\subfloat[$\mathcal{F}(\cdot)$]{\includegraphics[angle=90,width=1.in]{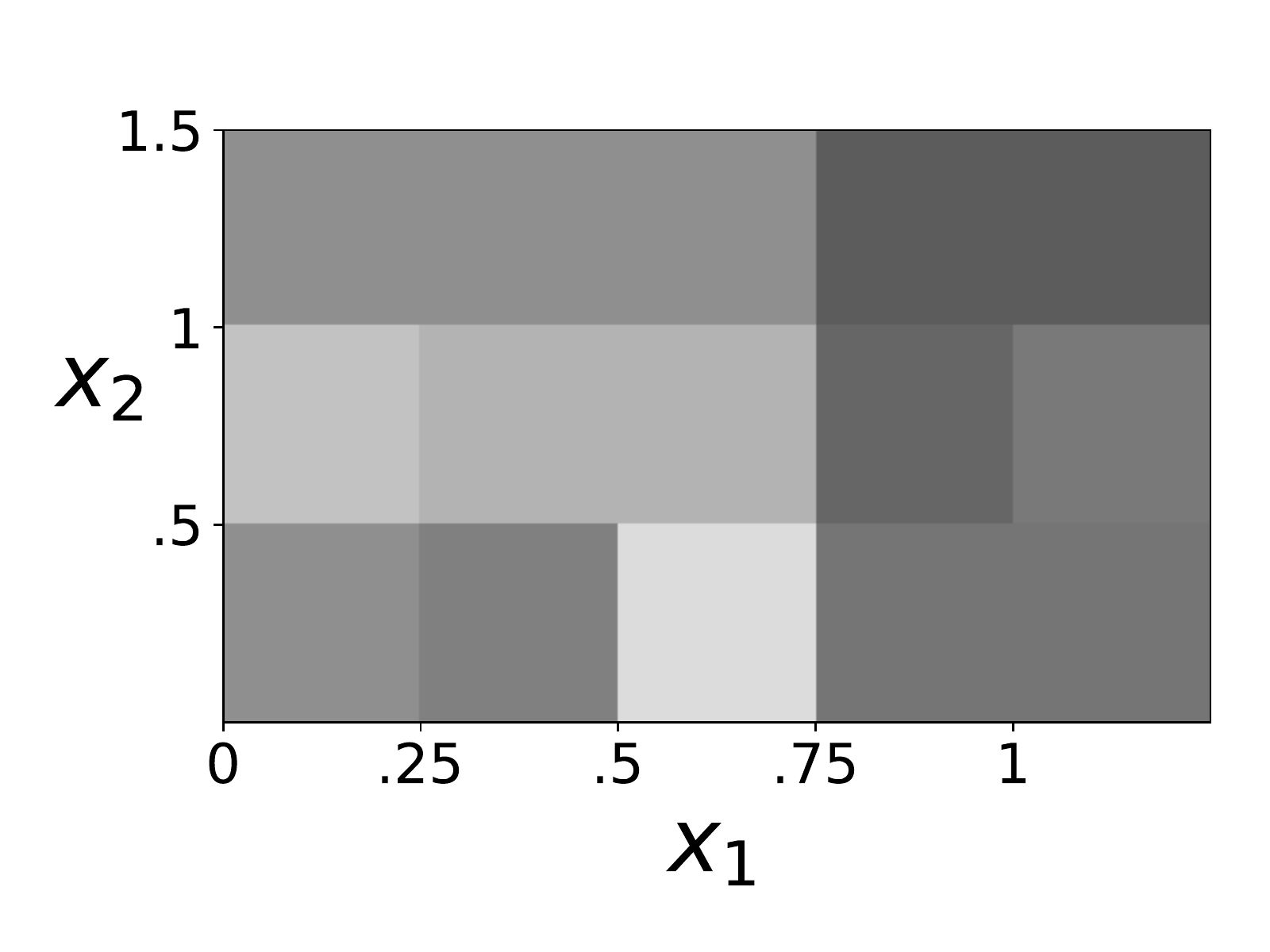}}
}
\caption{Example of a decision forest as a function of two variables $x_1$ and $x_2$. Intermediate nodes contain an axis-aligned decision and leaves, the filled nodes, return a numerical value. A visualization of the output of this function in 2-dimensional space is given in (b), where darker shades correspond to larger values. It is assumed that the output of the decision forest is an unweighted sum of the output of its constituent trees.}
\label{fig:decision_forest_function_heatmap}
\end{center}
\end{figure}

A \emph{decision forest} is a set of decision trees. The output of a decision forest is often an increasing monotonic function of the output of its individual decision trees, and varies depending on the task at hand. For regression, for example, it is a simple (weighted or unweighted) summation. In a Random Forest binary classifier, as another example, it is typically an $\argmax$ (i.e., majority vote). But to simplify exposition, we define the output to be an unweighted sum---our method is agnostic to and easily extends to other formulations as well. Then a decision forest $\mathcal{F}$ can be expressed as follows:
\begin{equation}
    \mathcal{F}(\bm{x}) \triangleq \sum_{\mathcal{T} \in \mathcal{F}} \sum_{R \in \mathcal{R_T}} \mathcal{L_T}(R) \mathbbm{1}_{\bm{x} \in R}.
    \label{equ:decision_forest_function}
\end{equation}
One could take the view that a decision forest, given the formulation above, is itself a partitioning of the feature space into regions $\mathcal{R_F}$, where each region $R$ has a constant value $\mathcal{L_F}(R)$. Equation~(\ref{equ:decision_forest_function}) can then be summarized as follows:
\begin{equation}
    \mathcal{F}(\bm{x}) \triangleq \sum_{R \in \mathcal{R_F}} \mathcal{L_F}(R) \mathbbm{1}_{\bm{x} \in R}.
    \label{equ:decision_forest_function_refined}
\end{equation}
Figure~\ref{fig:decision_forest_function_heatmap} illustrates this formulation for an example decision forest consisting of a pair of decision trees.

From Equation~(\ref{equ:decision_forest_function_refined}) it is clear that the gradients of $\mathcal{F}(\cdot)$ are 0 almost everywhere, with non-differentiability at the boundaries between regions due to a sudden jump in function value. As noted earlier, this property makes decision forests incompatible with gradient-based optimization methods.

%% file: proposed_method.tex
\section{Proposed Method}\label{sec:proposed_method}
\allowdisplaybreaks

As stated in earlier sections, we are interested in a unified setup where a decision forest can guide the (a) learning or (b) fine-tuning of embeddings, and still remain a proper decision forest in conformance with the definition in Section~\ref{sec:preliminaries}. In this section, we provide a detailed account of our proposal, starting with a description of the construction of our model, followed by the details of the training procedure, and closing with an analysis.

\subsection{Model Overview}
Let us assume there exists a function $\mathcal{E}:\mathbb{R}^m \rightarrow \mathbb{R}^d$ that projects an $m$-dimensional input into a $d$-dimensional embedding space. We require that $\mathcal{E}(\cdot)$ be differentiable in each dimension but otherwise do not impose any constraints. For example, $\mathcal{E}(\cdot)$ may be the identity function, a linear function, a parameterized non-linear neural network, or a combination thereof. As will be clear shortly, this flexibility allows the decision forest to consume tabular data directly (i.e., with an identity transformation) and raw structured data indirectly (i.e., as learnt embeddings).

Given $\mathcal{E}(\cdot)$, we now consider the composition $\mathcal{F}\circ\mathcal{E}(\cdot)$ where $\mathcal{F}$ is a decision forest as described in Equation~(\ref{equ:decision_forest_function_refined}). Where embeddings must be trained from scratch (i.e., when $\mathcal{E}$ is not pre-trained), $\mathcal{F}$ is simply randomly initialized: An input feature and a threshold in the interval $[0, 1]$ are randomly selected to form intermediate nodes, and the leaves are assigned random values (e.g., $0$ or $1$ for binary classification)---as we will explain shortly, leaves become trainable parameters of the model. We note that, we leave an examination of various techniques to randomly initialize decision forests to a future study. On the other hand, where $\mathcal{E}$ is pre-trained, $\mathcal{F}$ is a Random Forest, a GBDT, or any other type of decision forest trained on the output of $\mathcal{E}$.

That concludes the construction of our model. Completing a ``forward'' pass (i.e., taking an input example and producing a prediction) is trivial. But we have not yet addressed how this model is trained end-to-end. We will do just that next.

\subsection{Training}
As we are interested in the supervised learning setting, we assume we are given a \emph{loss} function, $\ell$, that is appropriate for the task at hand. Our only assumption about $\ell$ is that its gradient is available to us. The empirical risk minimization problem can then be expressed as minimizing the following objective:
\begin{equation}
    \frac{1}{|\Psi|}\sum_{(\bm{x}, y) \in \Psi} \ell(y, \mathcal{F}\circ\mathcal{E}(\bm{x})),
    \label{equ:optimization_problem}
\end{equation}
where $\Psi$ is the training dataset consisting of examples $\bm{x} \in \mathbb{R}^m$ and labels $y \in \mathbb{R}$, or in case of ranking, vectors of such pairs.

We are interested in minimizing Equation~(\ref{equ:optimization_problem}) using gradient descent. However, the decision forest $\mathcal{F}$ in our construction is either flat or discontinuous, and therein lies the obvious challenge. Our approach to addressing that challenge unsurprisingly involves a form of smoothing.

We approach this problem by smoothing not the structure of the decision forest---as is done in prior work---but by perturbing its input, $\bm{z} \in \mathbb{R}^d$. That is, instead of an input point falling into a single region of the decision forest as in Section~\ref{sec:preliminaries}, we allow it to stochastically belong to all regions. To that end, we assume that an input point $\bm{z}$ is itself the mean of a standard Gaussian distribution. We note that any symmetric distribution whose marginals can be decomposed into independent distributions is appropriate. We choose a Gaussian distribution simply because it is rather convenient for our analysis as its marginals are themselves also Gaussian distributions.

We have just injected uncertainty to the input of the decision forest. With the input perturbed, $\mathcal{F}$ is no longer determined by its value in a single region $R \in \mathcal{R_F}$, but by an expectation over all regions. The optimization problem can consequently be redefined as follows:
\begin{equation}
    \frac{1}{|\Psi|}\sum_{(\bm{x}, y) \in \Psi}
    \ell(y, 
    \mathop{\mathbb{E}}_{z \sim \mathcal{N}(\mu=\mathcal{E}(\bm{x}), \Sigma=\sigma\mathcal{I})}
    [ \mathcal{F}(\bm{z}) ]
    ),
    \label{equ:optimization_problem_redefined}
\end{equation}
where $\mathcal{I}$ is the identity matrix. It is clear that in the limit, when $\sigma$ approaches $0$ from above, Equation~(\ref{equ:optimization_problem_redefined}) approaches the original objective in Equation~(\ref{equ:optimization_problem}).

Let us now expand the expectation in Equation~(\ref{equ:optimization_problem_redefined}) and define the following:
\begin{equation}
    \widetilde{\mathcal{F}}_\sigma\circ\mathcal{E}(\bm{x}) \triangleq \mathop{\mathbb{E}}_{z \sim \mathcal{N}(\mathcal{E}(\bm{x}), \sigma\mathcal{I})}
    [ \mathcal{F}(\bm{z}) ]
    = \int_{\mathcal{R_F}} \mathcal{F}(z) f(z) dz,
\end{equation}
with $f(\cdot)$ denoting the probability density function of a Gaussian distribution. Using Equation~(\ref{equ:decision_forest_function_refined}), we arrive at the following:
\begin{equation}
    \widetilde{\mathcal{F}}_\sigma\circ\mathcal{E}(\bm{x}) = \sum_{R \in \mathcal{R_F}} \int_{R} \mathcal{F}(z) f(z) dz
    = \sum_{R \in \mathcal{R_F}} \mathcal{L_F}(R) \int_{R} f(z) dz.
    \label{equ:smooth_decision_forest}
\end{equation}

It is easy to calculate the gradients of Equation~(\ref{equ:smooth_decision_forest}) as, in each dimension, the integral is simply a difference in the cumulative distribution of the Gaussian. In fact, the integral decomposes into a product of $d$ independent terms where each term is a difference of the cumulative distribution function of a univariate Gaussian. This product is generally very sparse (with many of its terms being $1$) as not all features participate in forming every region. Such a decomposition makes the computation of gradients fast and scalable. For conciseness, we do not include the derivation of the gradients and, in practice, use automatic differentiation~\cite{abadi2016tensorflow} to do the computation.

Note that $\widetilde{\mathcal{F}}_\sigma\circ\mathcal{E}$ is also differentiable with respect to the leaf values and regions. An interesting implication then is that the leaf values (i.e., $\mathcal{L_F}(\cdot)$) or regions (i.e., $\mathcal{R_F}$) may even be considered parameters of the model and may be learnt or tuned. This is, in fact, the approach we take when learning embeddings from scratch: In an end-to-end training of the model, the leaf values of the decision forest as well as the parameters of the embedding function $\mathcal{E}$ are learnt. Note that making the regions trainable would lead to potential changes to the structure of the decision forest, though unlike prior work, the decisions are still hard and axis-aligned.

Finally, it is worth noting that the training procedure imposes no new restrictions on the model, nor does it necessitate any change to the nature of decisions in the decision forest; $\mathcal{F}$ remains a proper decision forest. Furthermore, the training procedure does not in practice require a perturbation of the actual data (or rather the output of the embedding function). Instead, all that is required can be done and is in fact achieved analytically: When computing the gradients of the loss with respect to model parameters $\Theta$, instead of computing $\nabla_{\Theta}\mathcal{F}\circ\mathcal{E}$ which is nonexistent, we simply calculate $\nabla_{\Theta}\widetilde{\mathcal{F}}_\sigma\circ\mathcal{E}$.

\begin{figure}[t]
\begin{center}
\centerline{
\includegraphics[width=3in,height=1.3in]{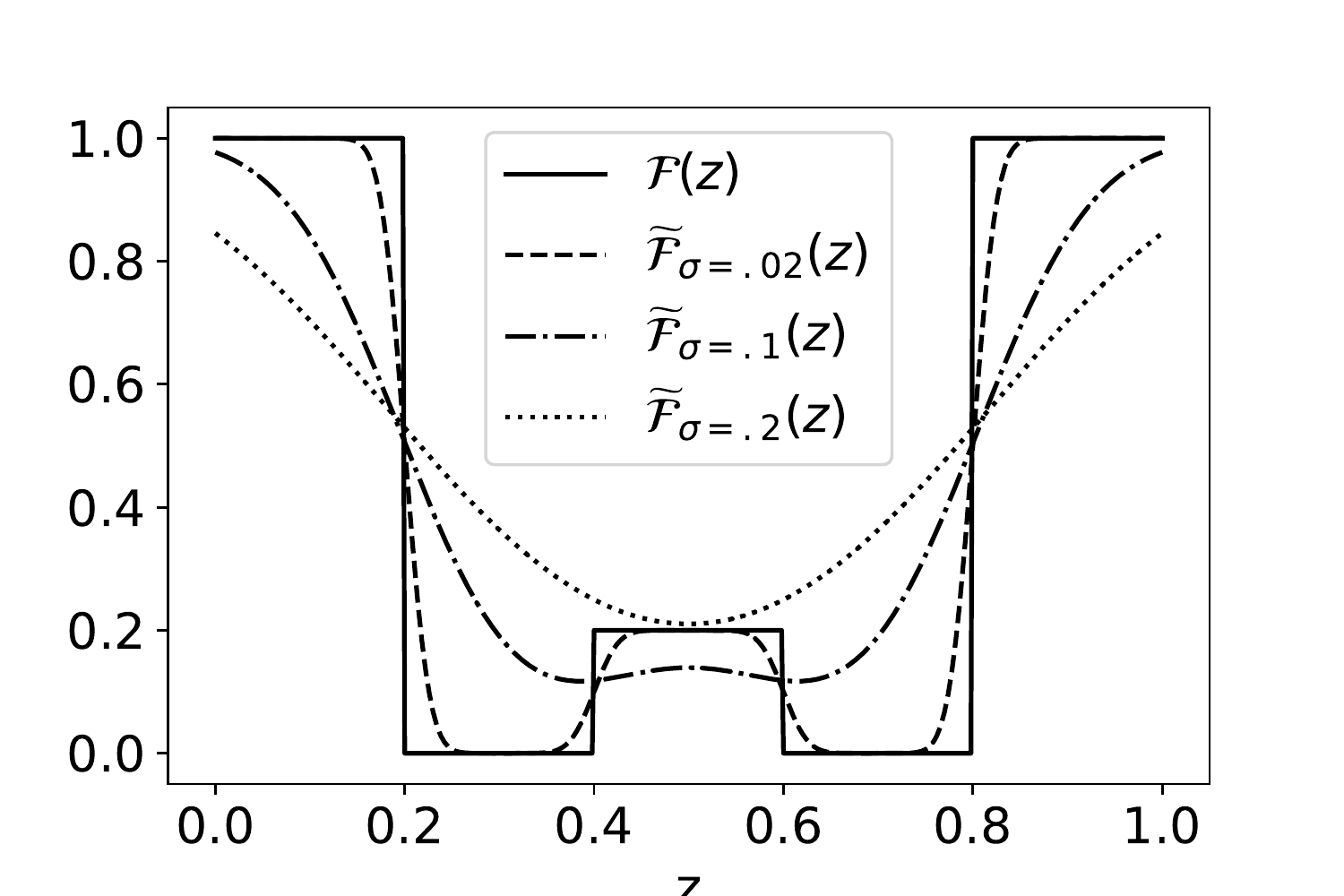}
}
\caption{The role of the Gaussian distribution's standard deviation $\sigma$ in the smoothing obtained by Equation~(\ref{equ:smooth_decision_forest}). The input $z$ is a point on the real line. The vertical axis represents the value of the decision forest $\mathcal{F}(z)$ and its smooth variant $\widetilde{\mathcal{F}}_\sigma(z)$ for various values of $\sigma$.}
\label{fig:smoothed_decision_forest_1d}
\end{center}
\end{figure}

\subsection{Analysis}
Consider an example in one dimension: A point $z \in \mathbb{R}$ and a decision tree $\mathcal{F}$ that simply partitions $\mathbb{R}$ into line segments. As $z$ sweeps the real line, the Gaussian distribution centered at $z$ allocates a different probability mass for each line segment. As a consequence, as $z$ approaches a decision boundary, the value of $\widetilde{\mathcal{F}}_\sigma(z)$ effectively becomes an interpolation of the value of $\mathcal{F}$ in adjacent segments. Changing the standard deviation $\sigma$ of the Gaussian noise adjusts the influence of distant segments, with very small values of $\sigma$ effectively limiting the smoothing effect only to areas close to decision boundaries. This phenomenon is depicted in Figure~\ref{fig:smoothed_decision_forest_1d}.

What we observe in one dimension extends naturally to multiple dimensions, a direct result of our choice of the noise distribution. Figure~\ref{fig:smoothed_decision_forest_2d} illustrates an example decision forest and its smoothed variants with different values for the standard deviation $\sigma$.

An interesting result of this behavior is that by changing the standard deviation of the underlying distribution, we are able to modify the optimization landscape. As $\sigma$ becomes smaller, $\widetilde{\mathcal{F}}_\sigma\circ\mathcal{E}$ morphs into what is effectively $\mathcal{F}\circ\mathcal{E}$ almost everywhere except at boundaries. In fact, one could begin with a larger $\sigma$ and gradually decrease its magnitude as training continues. We have experimented with this \emph{annealing} procedure, though found that it often leads to similar optima as the ones obtained by simply fixing $\sigma$ to a value that is carefully tuned on a validation set.

%% file: experiments.tex
\section{Experimental Evaluation}\label{sec:experiments}

Having described our proposed method in the previous section, we now turn to its empirical evaluation. This section provides a description of our experimental setup including the datasets we used in this work, and presents and discusses our findings.

Before we begin, let us reiterate the two research questions we attempt to investigate: Can we (RQ1) learn embeddings and (RQ2) fine-tune pre-trained embeddings using proper decision forests? It is worth noting that, our examination of these questions are agnostic to the choice of the embedding function $\mathcal{E}(\cdot)$: We are only interested in verifying the feasibility of our approach for RQ1, and any relative gains obtained by fine-tuning pre-trained embeddings for RQ2. In other words, choosing $\mathcal{E}$ differently is inconsequential for our investigation and does not invalidate any of our assertions. Therefore, in our experiments, $\mathcal{E}(\cdot)$ is either a simple feed-forward neural network or an off-the-shelf pre-trained module. Though, in practice, it is easy to replace $\mathcal{E}$ with the latest and most appropriate embedding function.

\begin{figure}[t]
\begin{center}
\centerline{
\subfloat[$\mathcal{F}(x_1, x_2)$]{\includegraphics[width=1.8in,height=.9in]{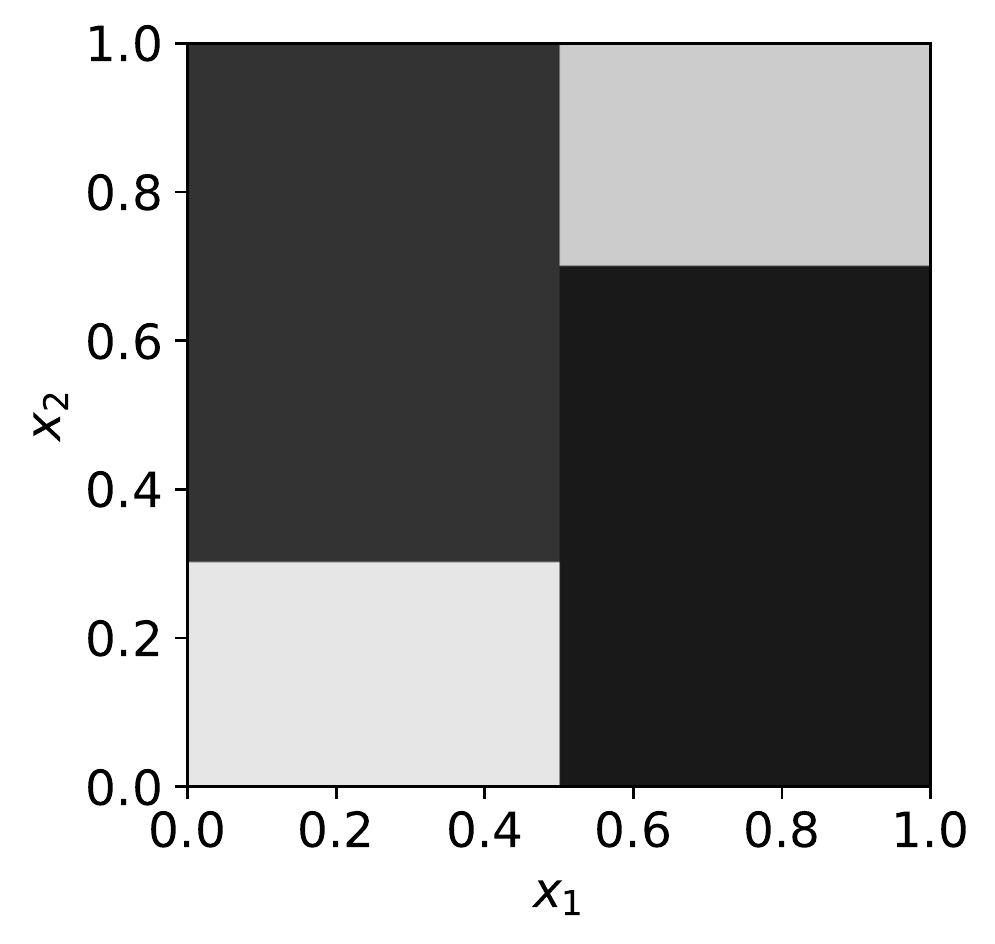}}
\subfloat[$\widetilde{\mathcal{F}}_{\sigma=.05}(x_1, x_2)$]{\includegraphics[width=1.8in,height=.9in]{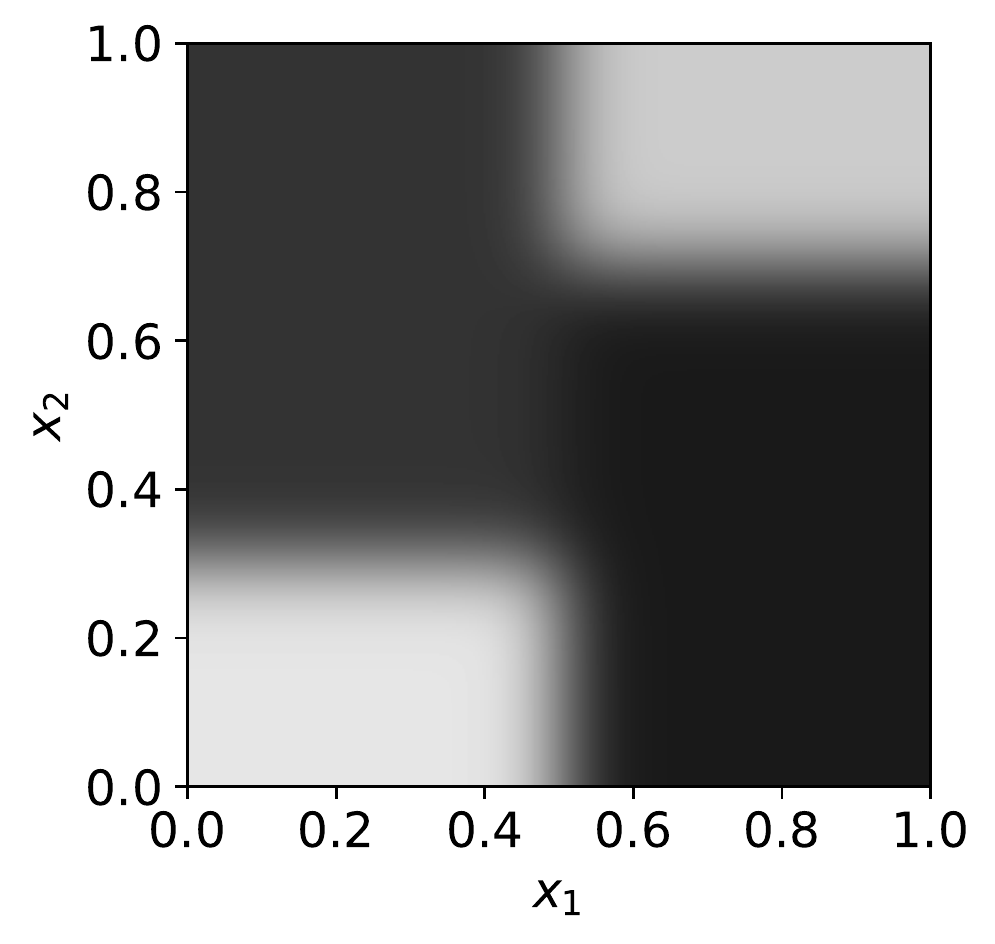}}
}
\vspace{-0.15in}
\centerline{
\subfloat[$\widetilde{\mathcal{F}}_{\sigma=.10}(x_1, x_2)$]{\includegraphics[width=1.8in,height=.9in]{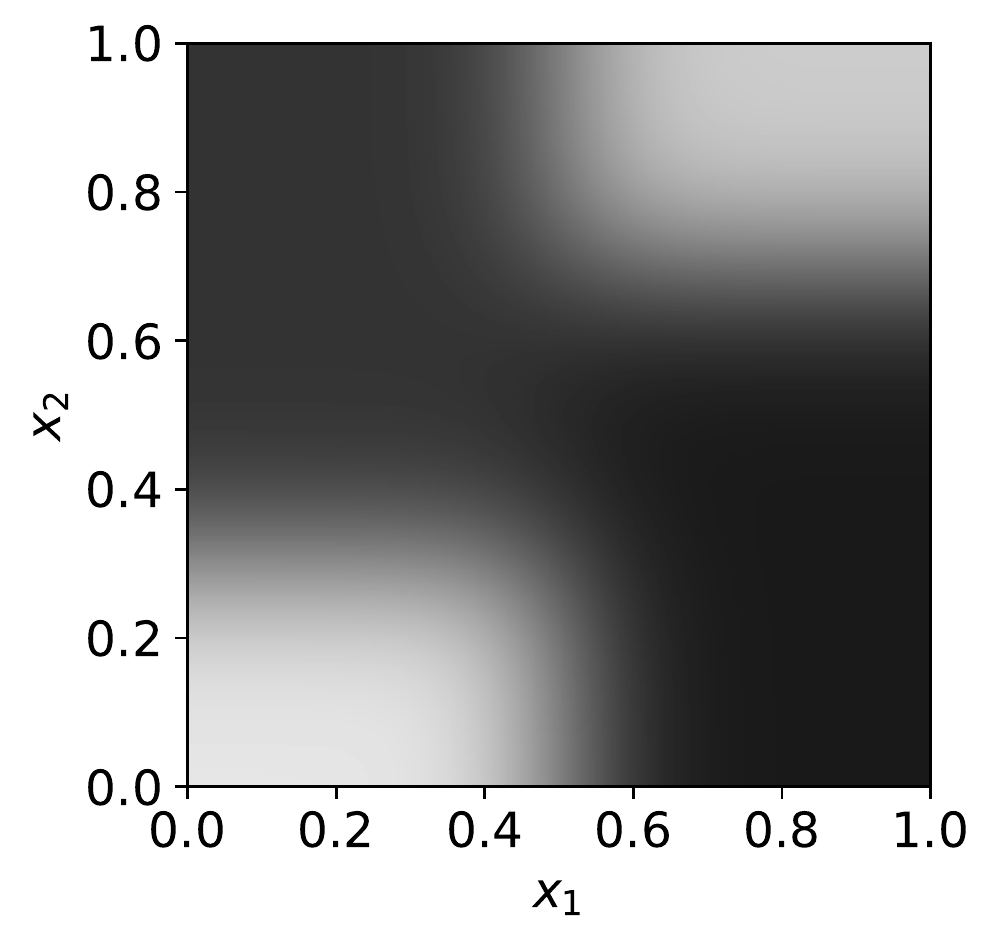}}
\subfloat[$\widetilde{\mathcal{F}}_{\sigma=.15}(x_1, x_2)$]{\includegraphics[width=1.8in,height=.9in]{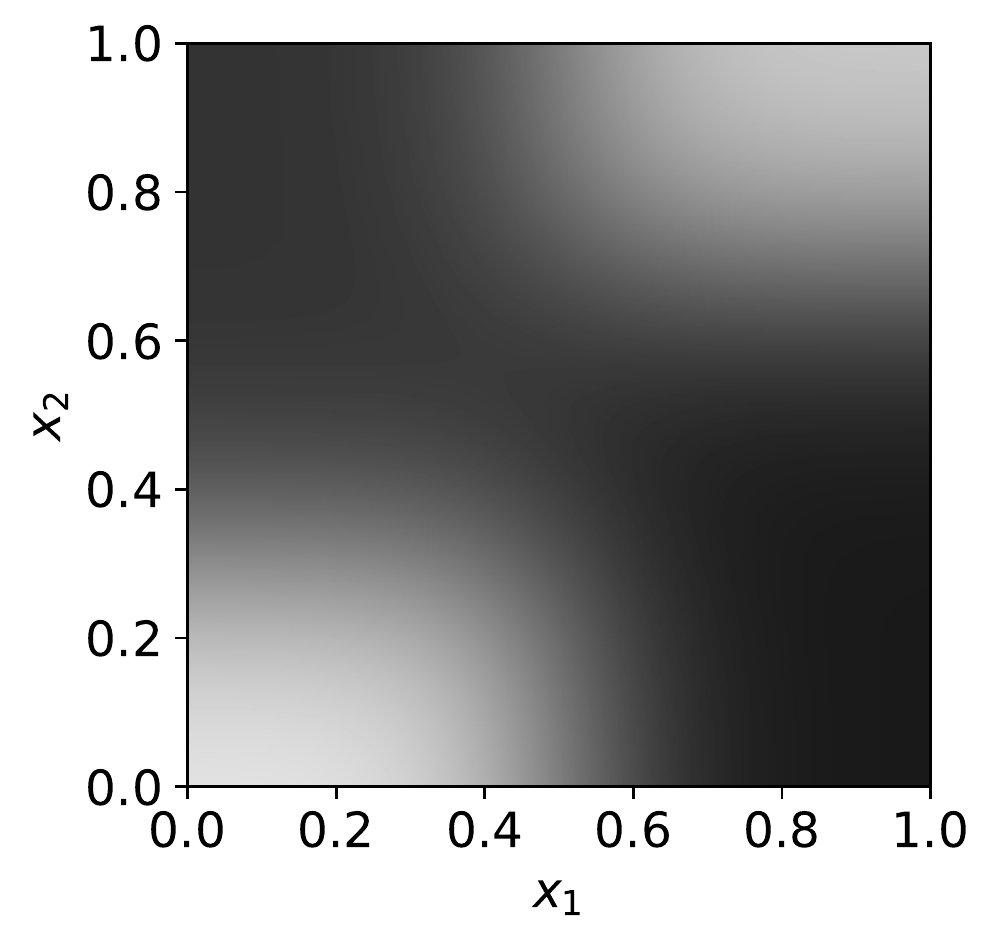}}
}
\caption{The effect of the Gaussian distribution's standard deviation $\sigma$ on the smoothing of the decision forest $\mathcal{F}$ in 2-dimensional space using Equation~(\ref{equ:smooth_decision_forest}). Axes represent individual features and the different shades correspond to different leaf values.}
\label{fig:smoothed_decision_forest_2d}
\end{center}
\end{figure}

\subsection{Experimental Setup}
We have implemented the proposed method, which we intend to open-source, in Tensorflow~\cite{abadi2016tensorflow}. Our choice of Tensorflow was motivated by (a) the framework's simplification of gradient computation through automatic differentiation, and (b) the availability of state-of-the-art pre-trained encoders as well as a multitude of real-world benchmark datasets through Tensorflow Hub~\footnote{Available at http://tfhub.dev with code at http://github.com/tensorflow/hub}.

In investigating RQ1, the embedding function $\mathcal{E}$ is a feed-forward neural network with $d$ output neurons, where we choose $d$ depending on the dataset. We will state the architecture of the neural network (i.e., number of hidden layers and count of output neurons, $d$) in upcoming sections. The neural network is randomly initialized.

As for the decision forest $\mathcal{F}$ in RQ1, it suffices to generate a set of random decision trees. As explained in Section~\ref{sec:proposed_method}, to generate a decision tree randomly, we follow a recursive procedure: We choose a feature at random and sample a threshold uniformly randomly from the interval $[0, 1]$ to form an intermediate node, and repeat this process for the left and right sub-trees, until a depth limit is reached. The leaf nodes are initialized with random values appropriate for the task (e.g., $\{0, 1\}$ for binary classification). Again, we state the depth limit in the discussion of each experiment in upcoming sections. Finally, leaf values are added as trainable parameters of the model.

The setup for RQ2 is similar, but we do not initialize $\mathcal{E}$ and $\mathcal{F}$ randomly. Instead, as $\mathcal{E}$, we use the Universal Sentence Encoder~\cite{cer2018universal}---a pre-trained encoder available in Tensorflow Hub and appropriate for text classification and natural language processing tasks. To form $\mathcal{F}$, we train a Random Forest or a GBDT on the output of $\mathcal{E}(\cdot)$. Unlike in RQ1 where the leaves of the decision forest are trainable, the decision forest in RQ2 does not have any trainable parameters; only $\mathcal{E}$ is expected to be fine-tuned.

As for the tree training algorithm, one is free to use any proprietary or open-source library such as LightGBM~\cite{lightgbm2017nips}, XGBoost~\cite{Chen2016XGBoost}, or Scikit-Learn~\cite{Pedregosa2011ScikitLearn}. We will include code to consume the more common representation of decision forests and convert those to our data structure in Tensorflow. In the experiments below, we use XGBoost to train GBDTs on pre-trained embeddings.

\subsection{Synthetic Datasets}
In a first set of experiments, we examine RQ1 using synthetic datasets. Evaluation with synthetic datasets allows us to design patterns that are difficult to model with a decision forest alone, and makes it possible to visually inspect the output of the model. In what follows, we describe how we generate these datasets and discuss our findings.

\begin{figure}
\begin{center}
\centerline{
\subfloat[Dataset]{\includegraphics[width=1.15in,height=1.in]{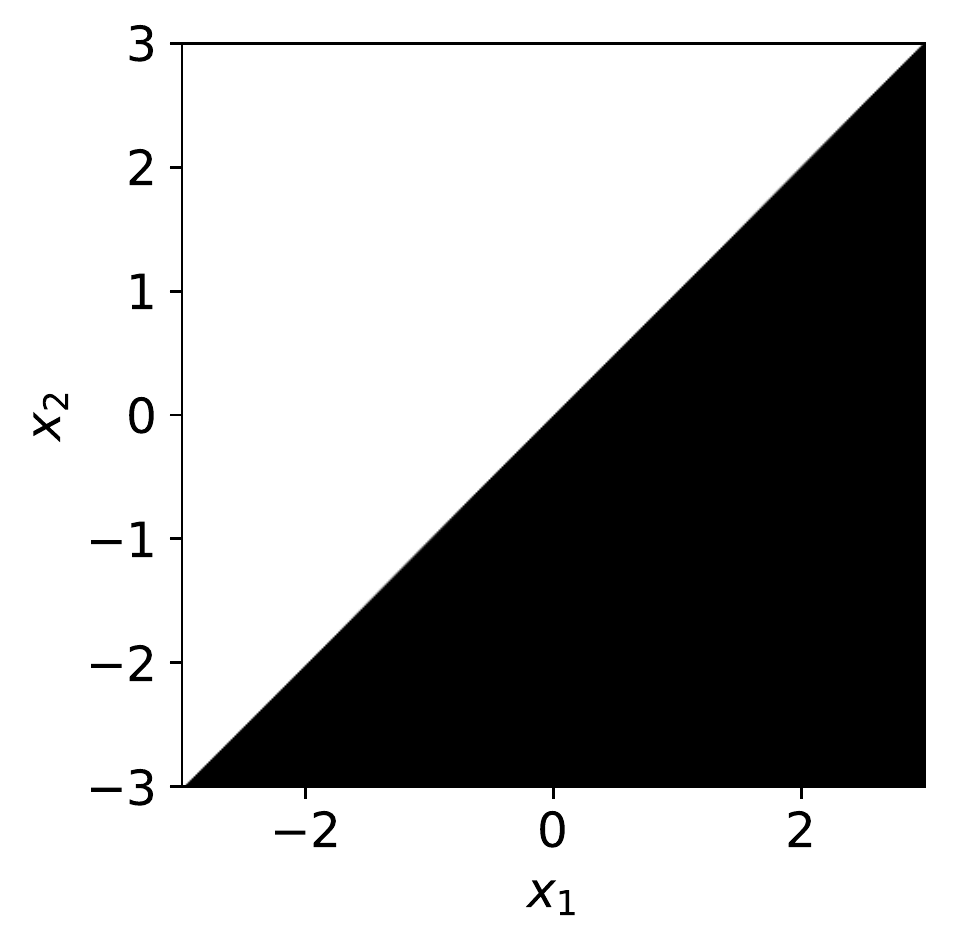}}
\subfloat[Initial $\mathcal{E}(\cdot)$]{\includegraphics[width=1.15in,height=1.in]{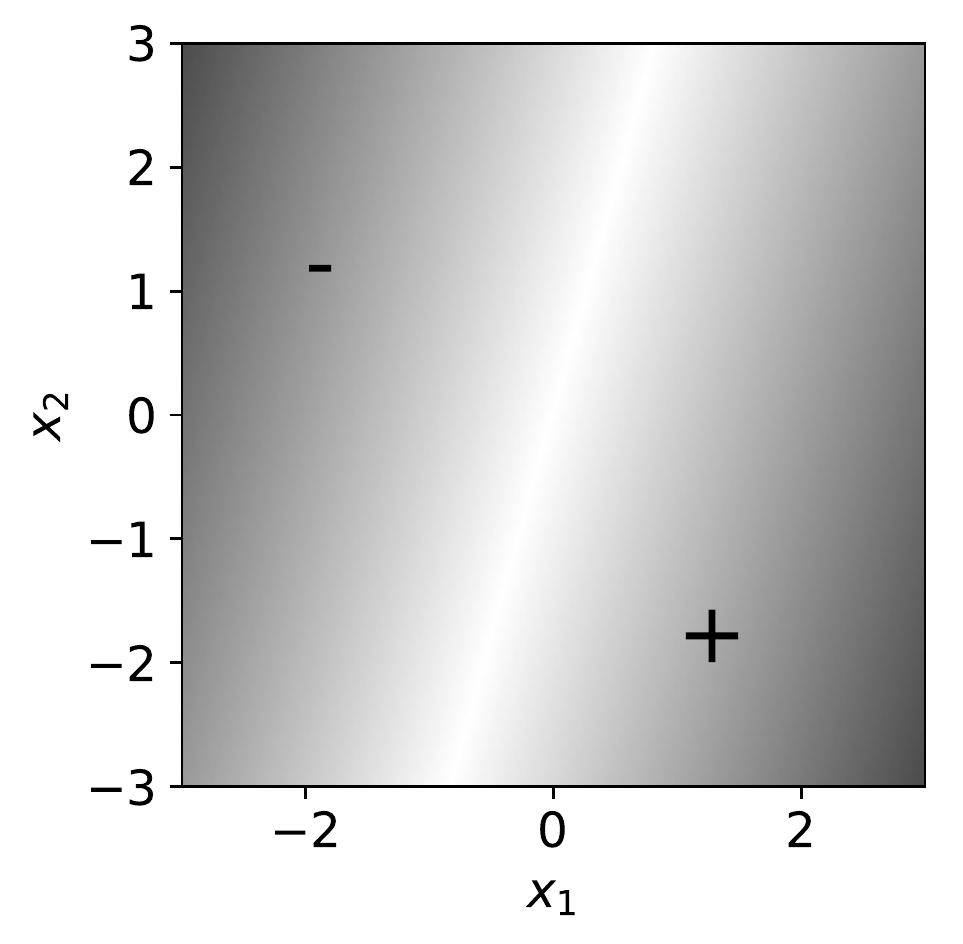}}
\subfloat[Trained $\mathcal{E}(\cdot)$]{\includegraphics[width=1.15in,height=1.in]{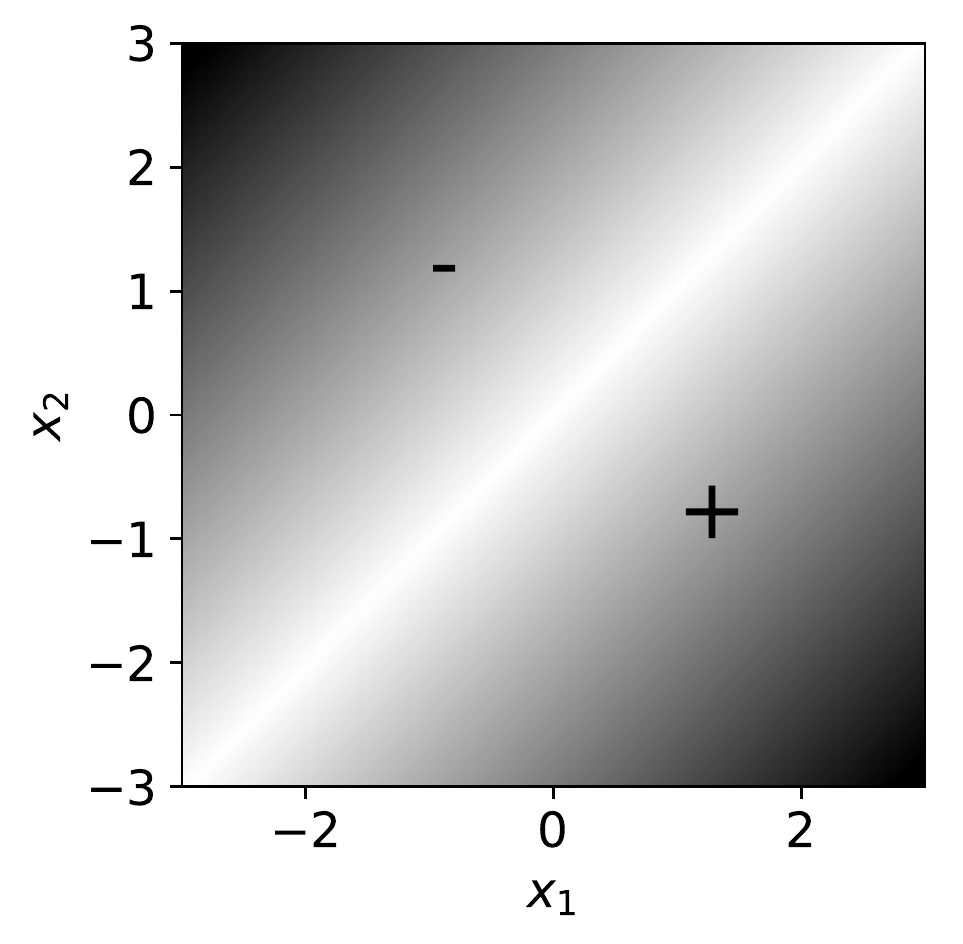}}
}
\caption{Binary classification problem with a synthetic dataset in $\mathbb{R}^2$ where a point's label is 1 if $x_1 > x_2$ and 0 otherwise. The dataset is shown in (a). (b) and (c) visualize the magnitude of the embedding function $\mathcal{E}(\cdot)$, which is a single neuron, before and after training---darker shades correspond to larger values, and the $+$ or $-$ annotations show the sign of the embedding function. It is clear that $\mathcal{E}$ has learnt to project $(x_1, x_2)$ to $\mathcal{E}(x_1, x_2) = x_1 - x_2$.}
\label{fig:synthetic_identity_line_decision_boundary}
\end{center}
\end{figure}

\subsubsection{Identity line as decision boundary}
Our very first experiment serves as a proof of concept. We consider a binary classification dataset in $\mathbb{R}^2$ whose positive and negative examples (i.e., points of the form $(x_1, x_2)$) are separated by the identity line $x_1 = x_2$. In other words, the label of the point $(x_1, x_2)$ is 1 if $x_1 > x_2$ and is 0 otherwise, as illustrated in Figure~\ref{fig:synthetic_identity_line_decision_boundary}(a).

This is a difficult decision boundary to model with a decision forest alone because, as we have already noted, splits in a decision forest are axis-aligned, leading to axis-aligned decision boundaries. Notably, axis-aligned decision boundaries do not generalize well to points in this dataset.

\begin{figure}
\begin{center}
\centerline{
\subfloat{\includegraphics[width=1.15in,height=1.in]{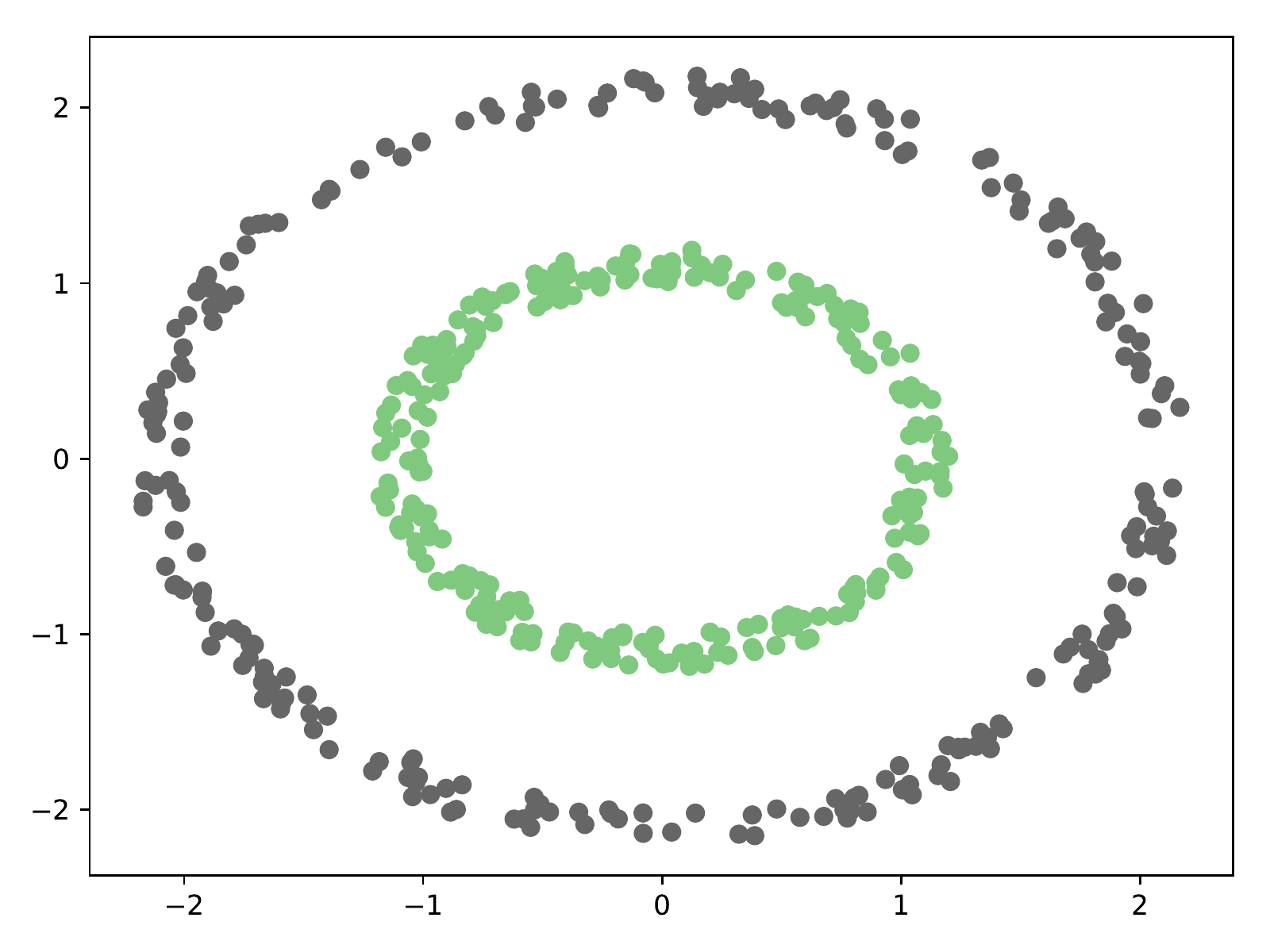}}
\subfloat{\includegraphics[width=1.15in,height=1.in]{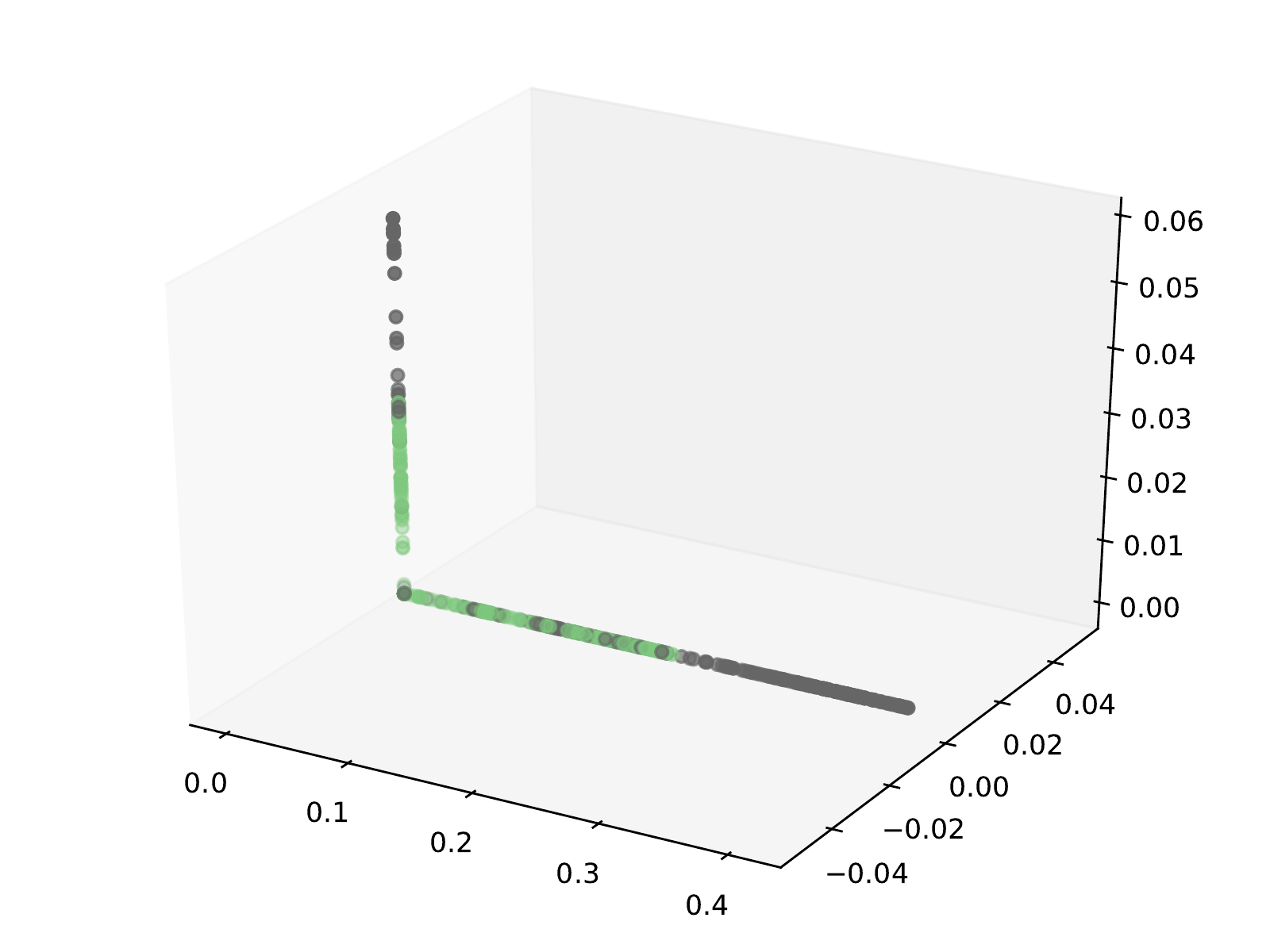}}
\subfloat{\includegraphics[width=1.15in,height=1.in]{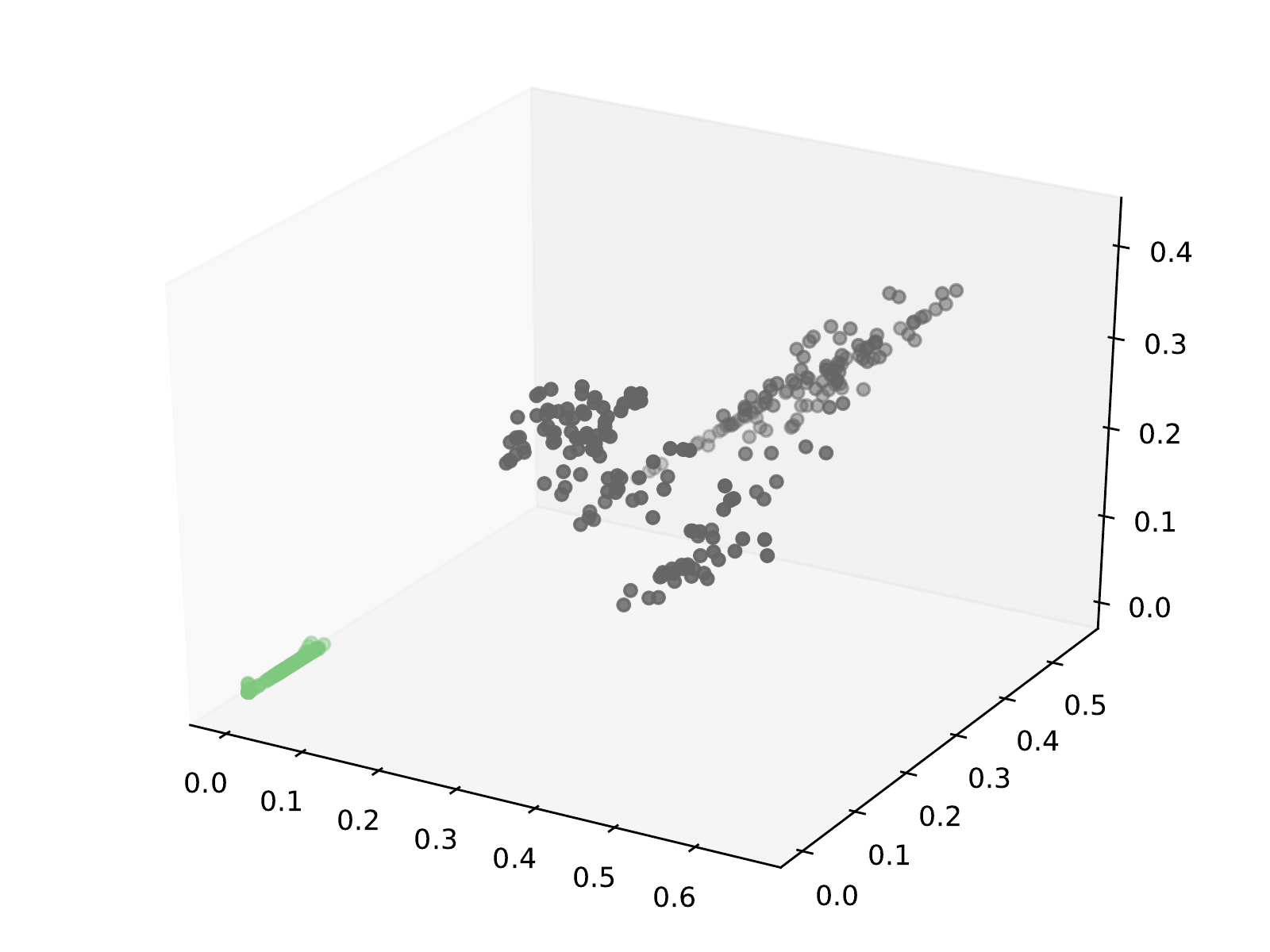}}
}

\centerline{
\subfloat{\includegraphics[width=1.15in,height=1.in]{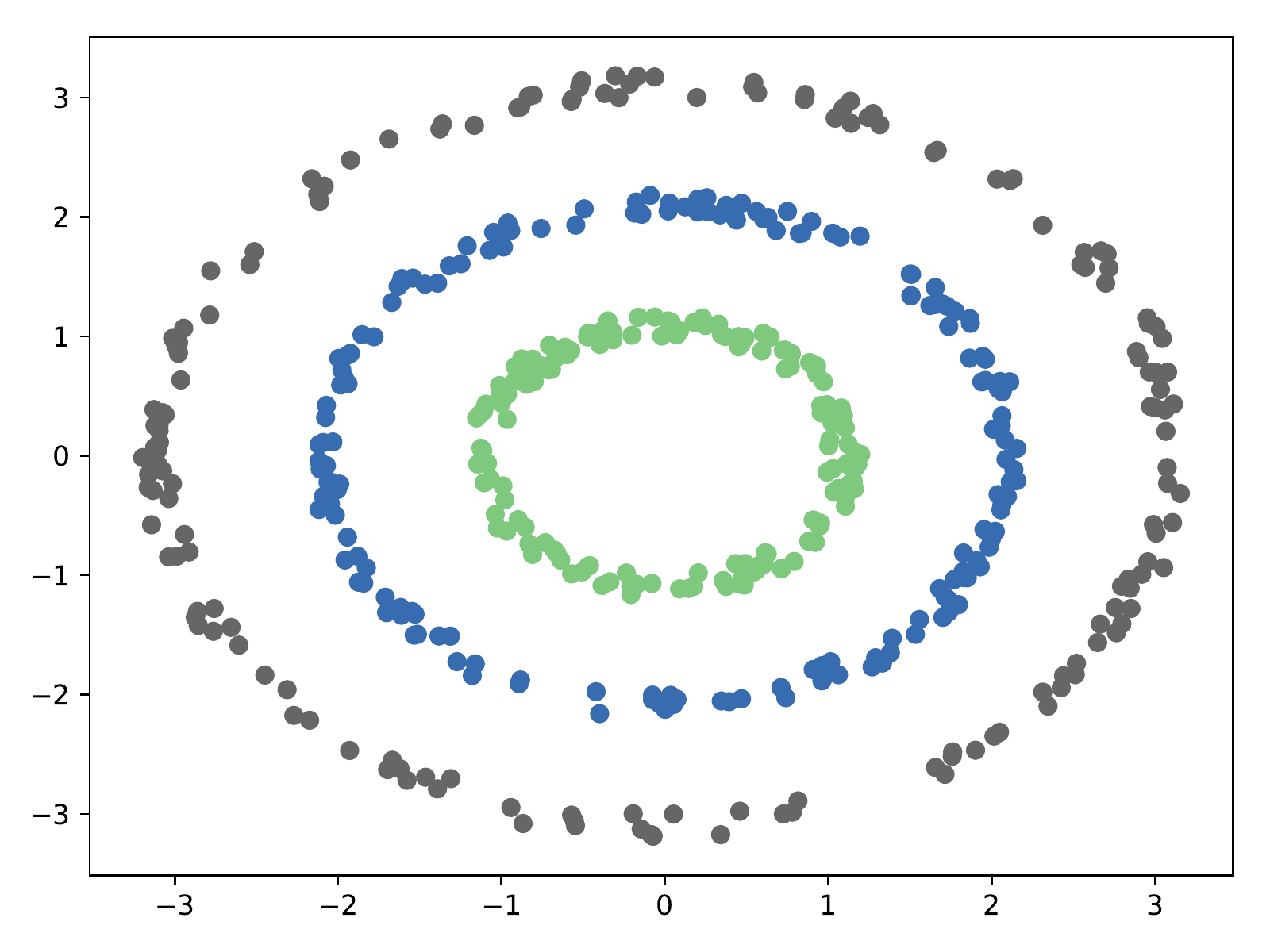}}
\subfloat{\includegraphics[width=1.15in,height=1.in]{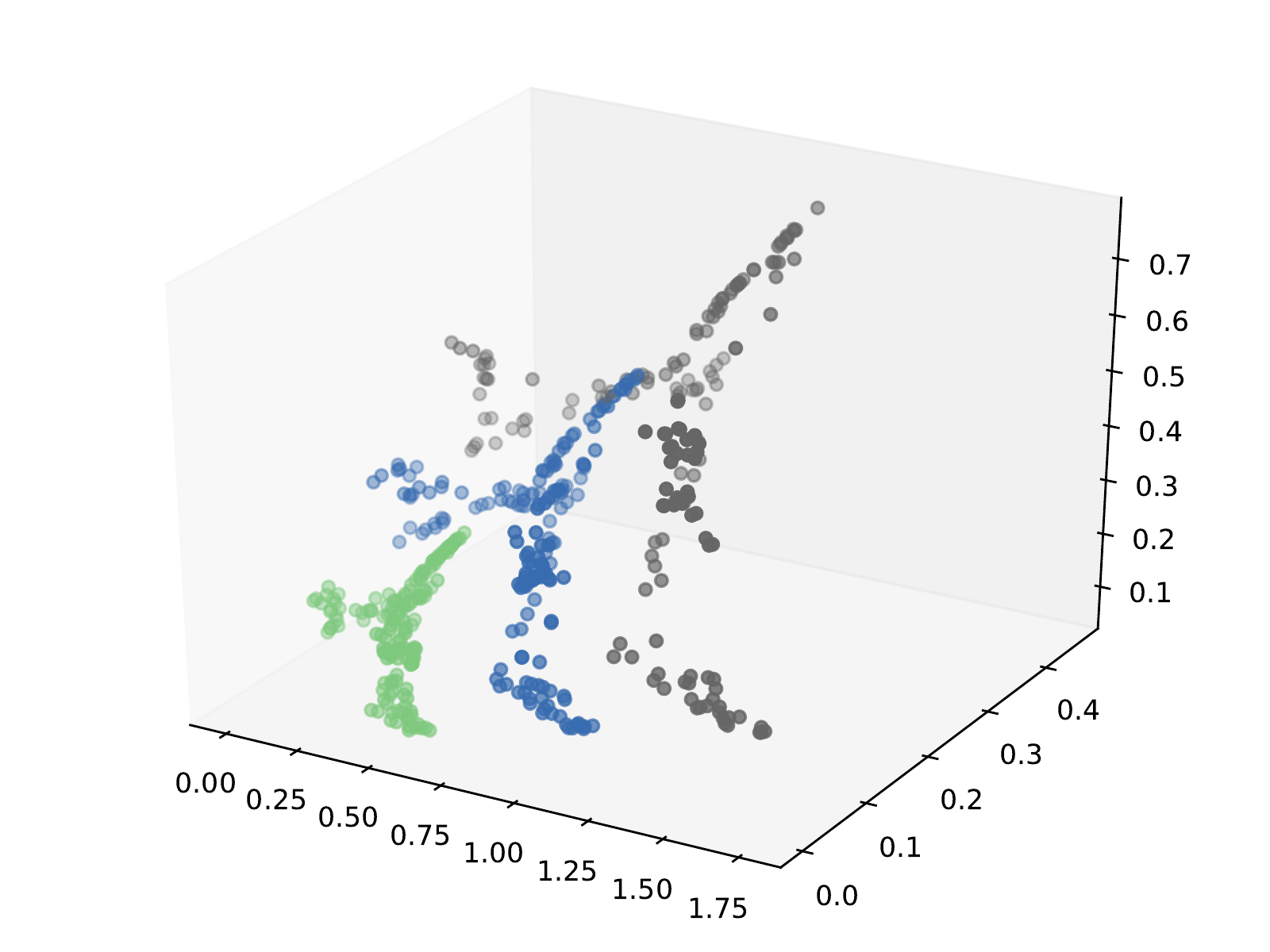}}
\subfloat{\includegraphics[width=1.15in,height=1.in]{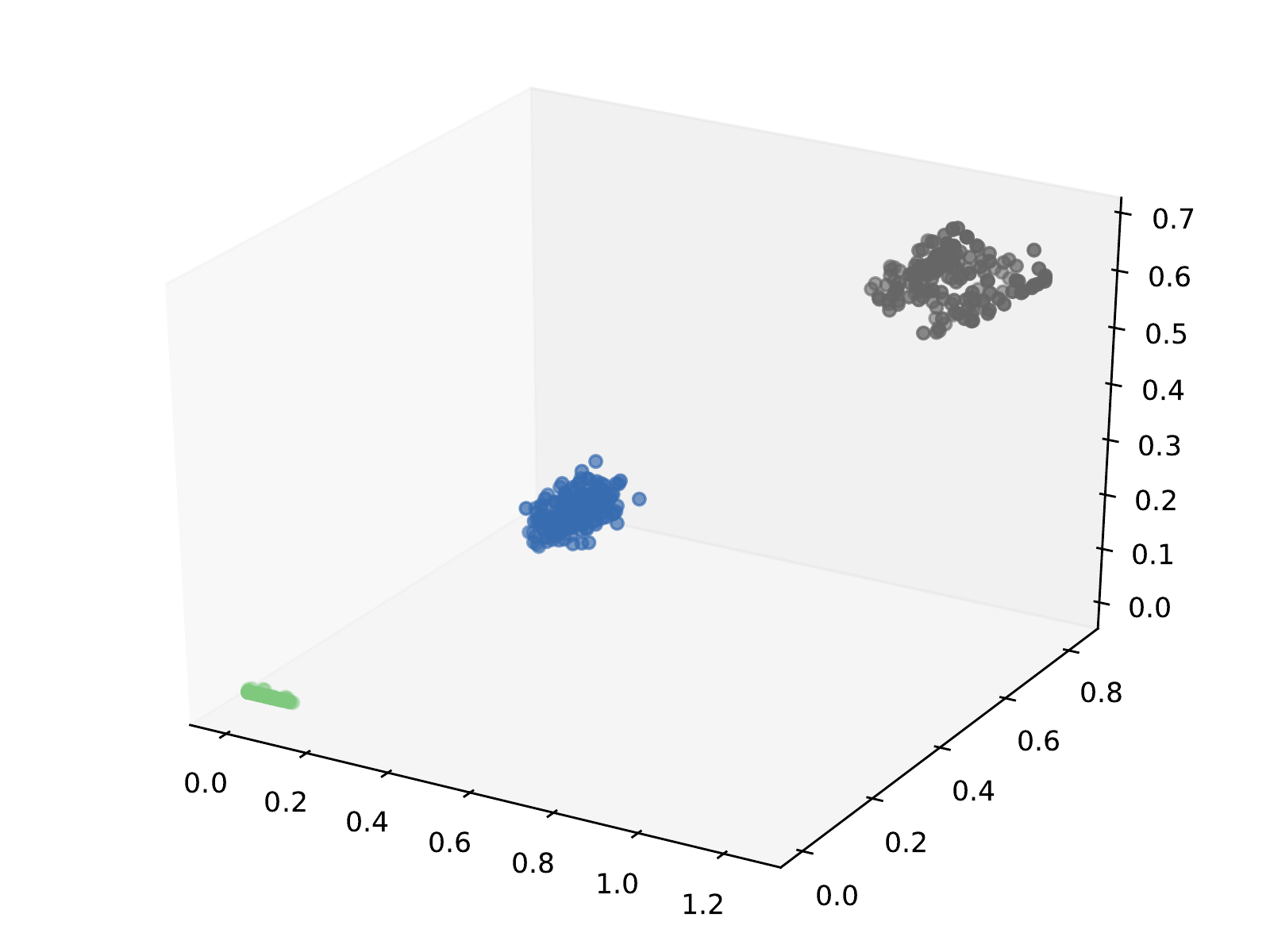}}
}

\centerline{
\subfloat{\includegraphics[width=1.15in,height=1.in]{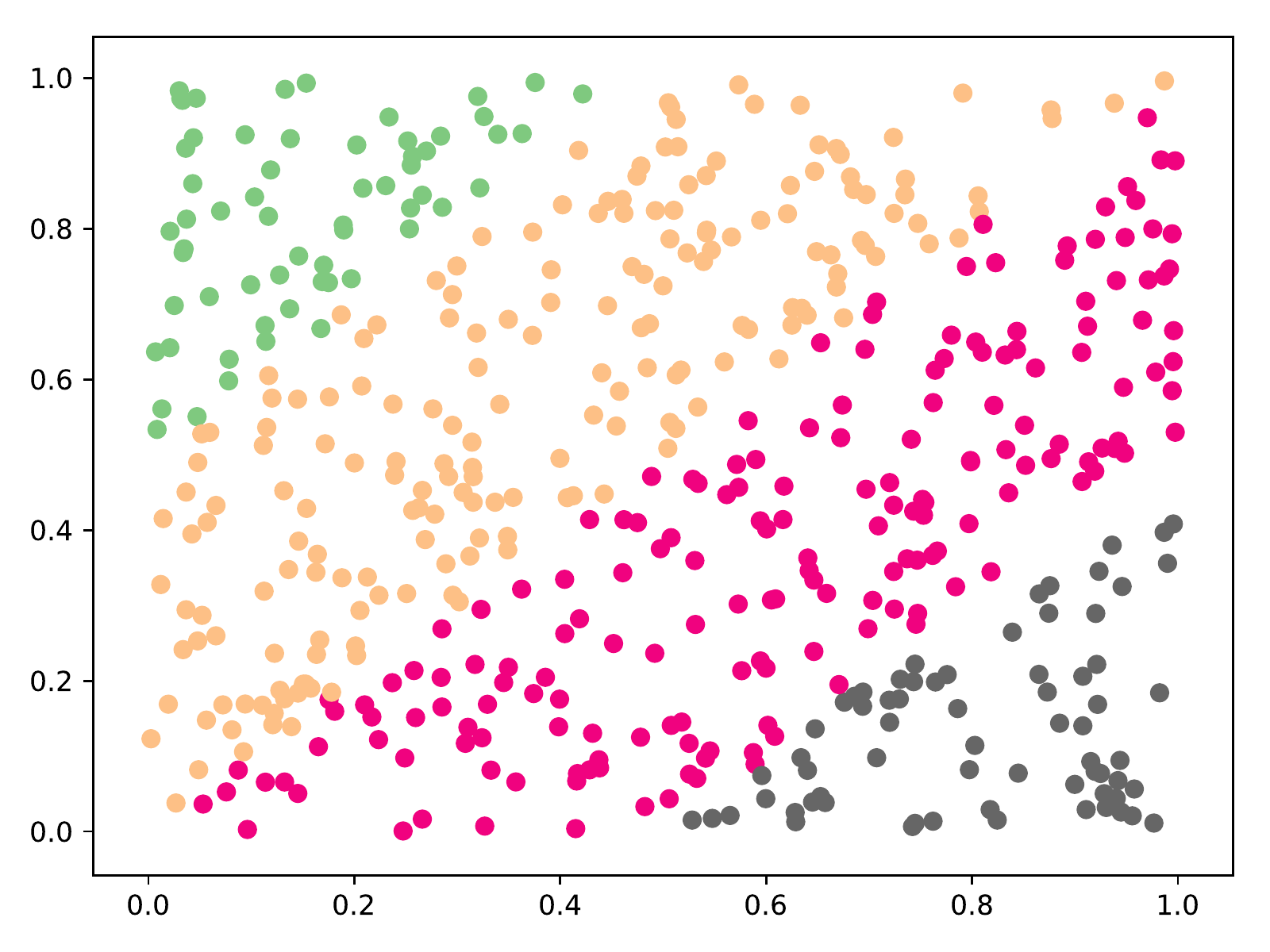}}
\subfloat{\includegraphics[width=1.15in,height=1.in]{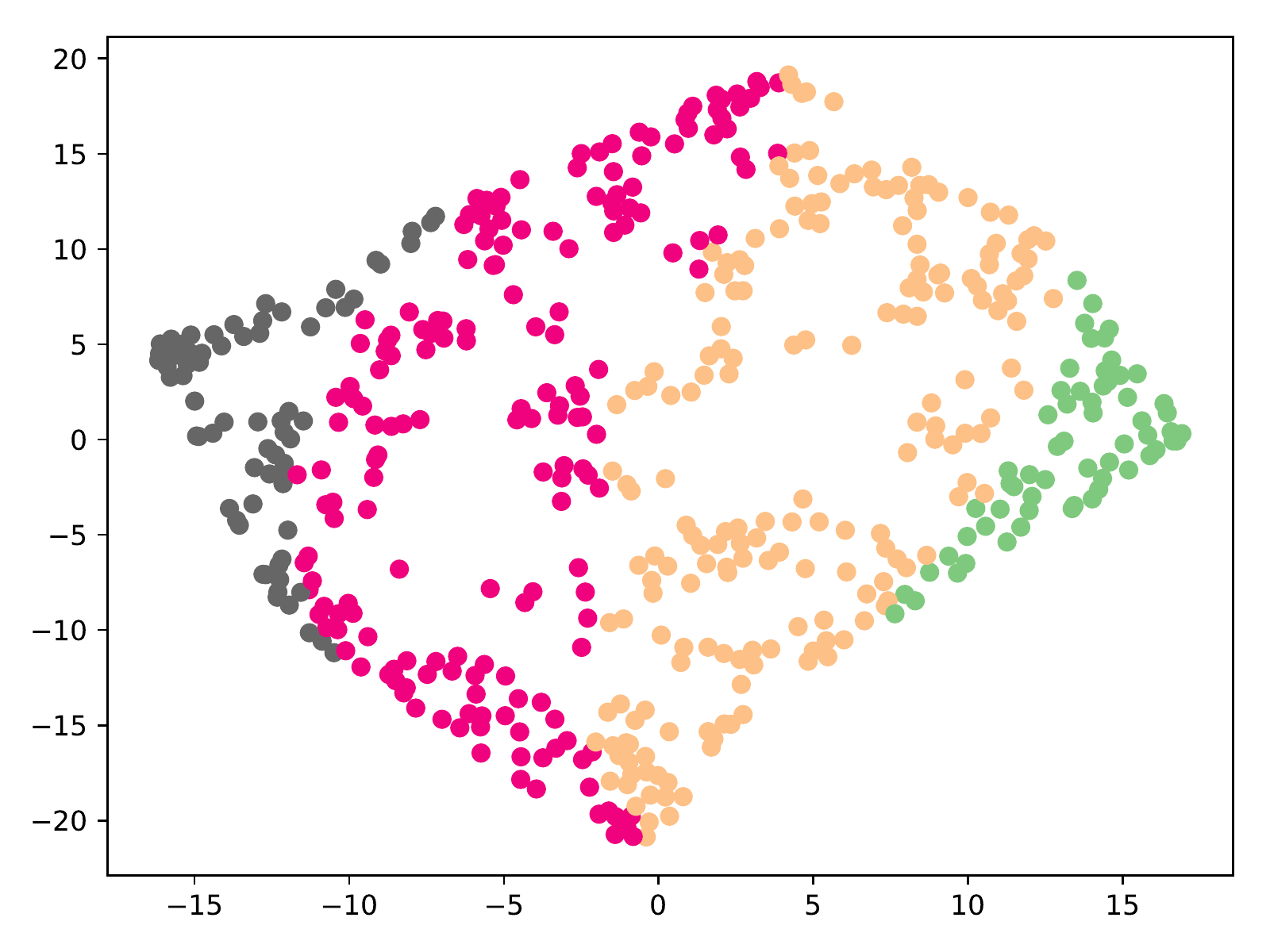}}
\subfloat{\includegraphics[width=1.15in,height=1.in]{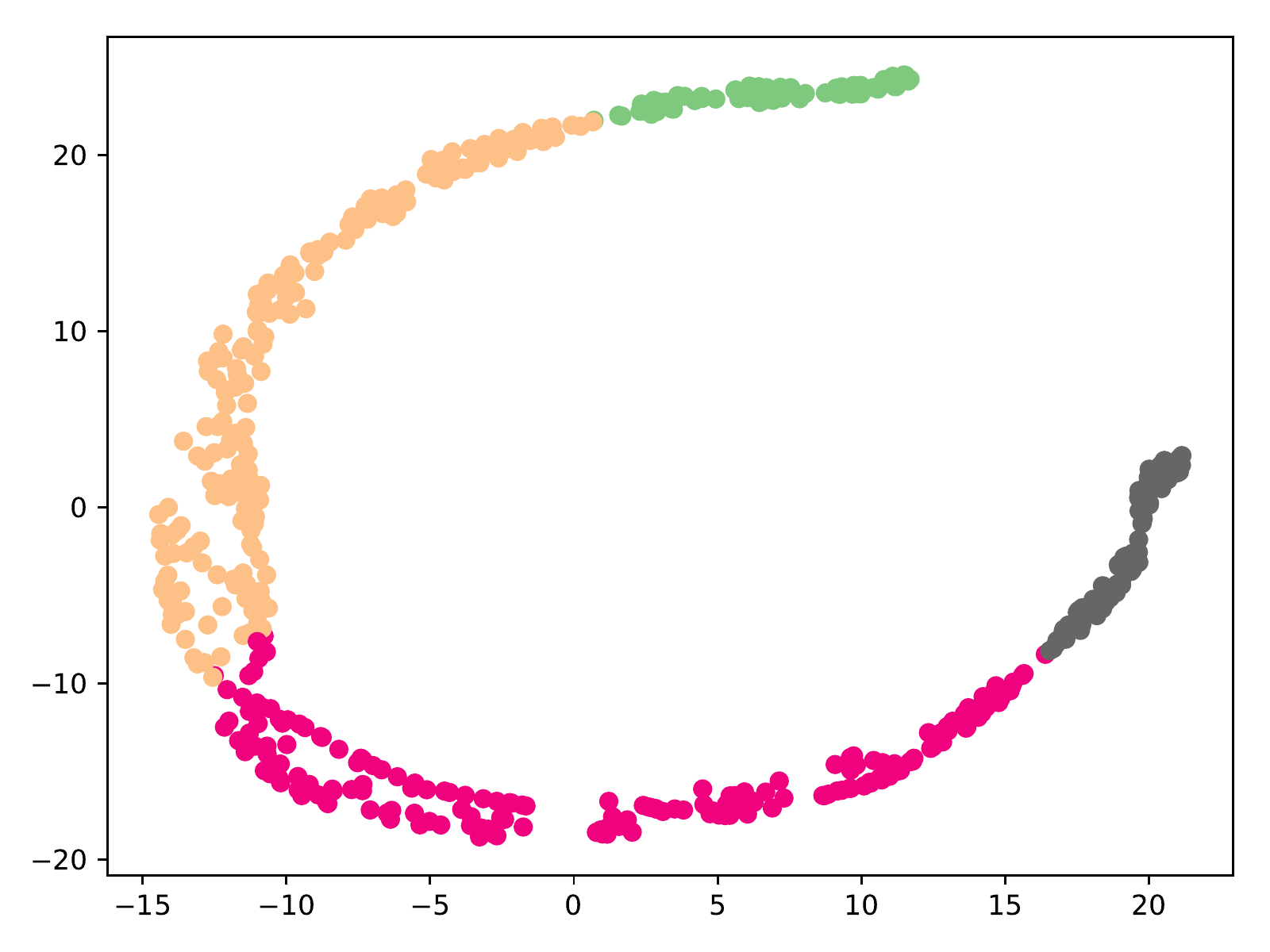}}
}

\centerline{
\subfloat{\includegraphics[width=1.15in,height=1.in]{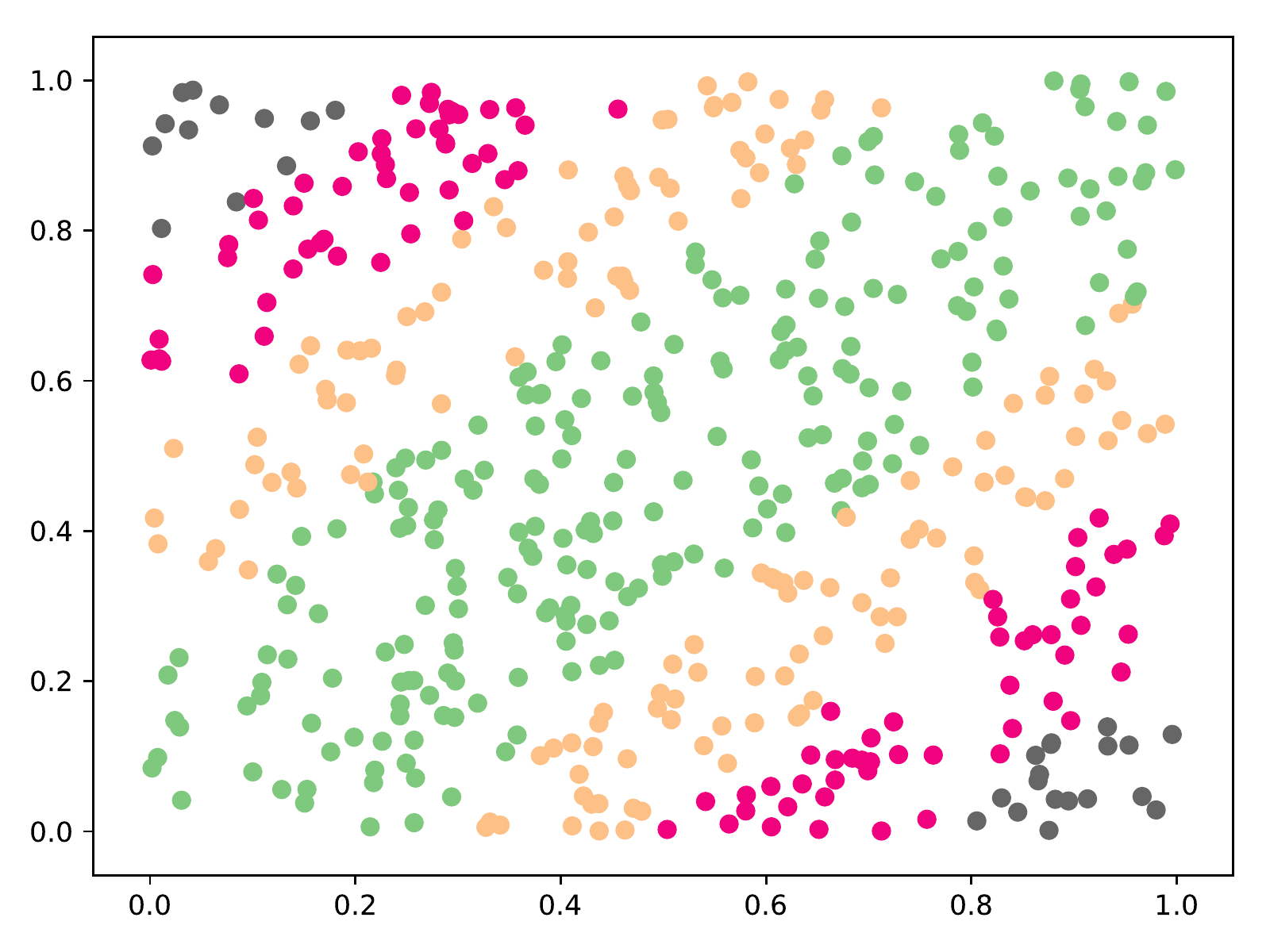}}
\subfloat{\includegraphics[width=1.15in,height=1.in]{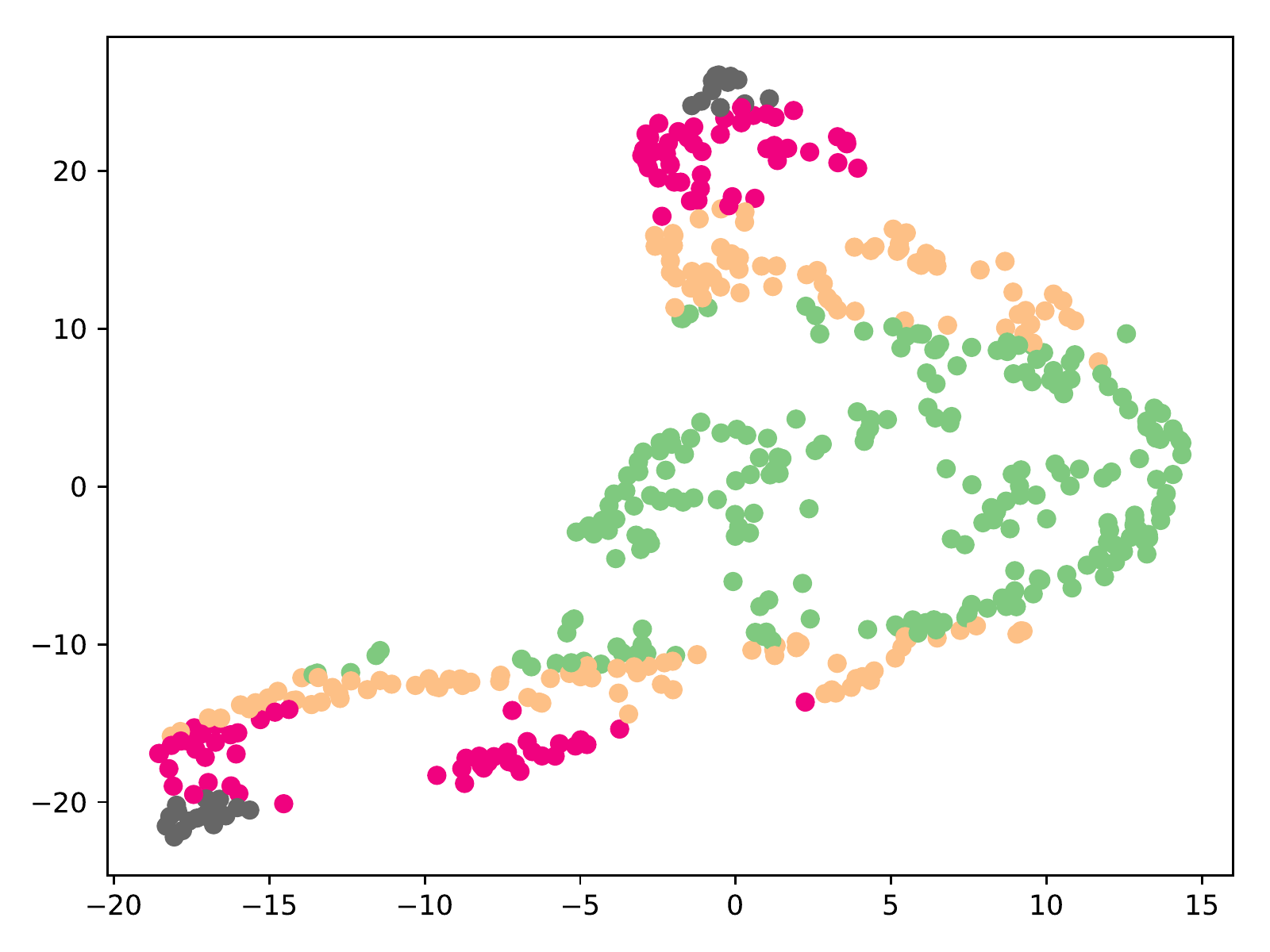}}
\subfloat{\includegraphics[width=1.15in,height=1.in]{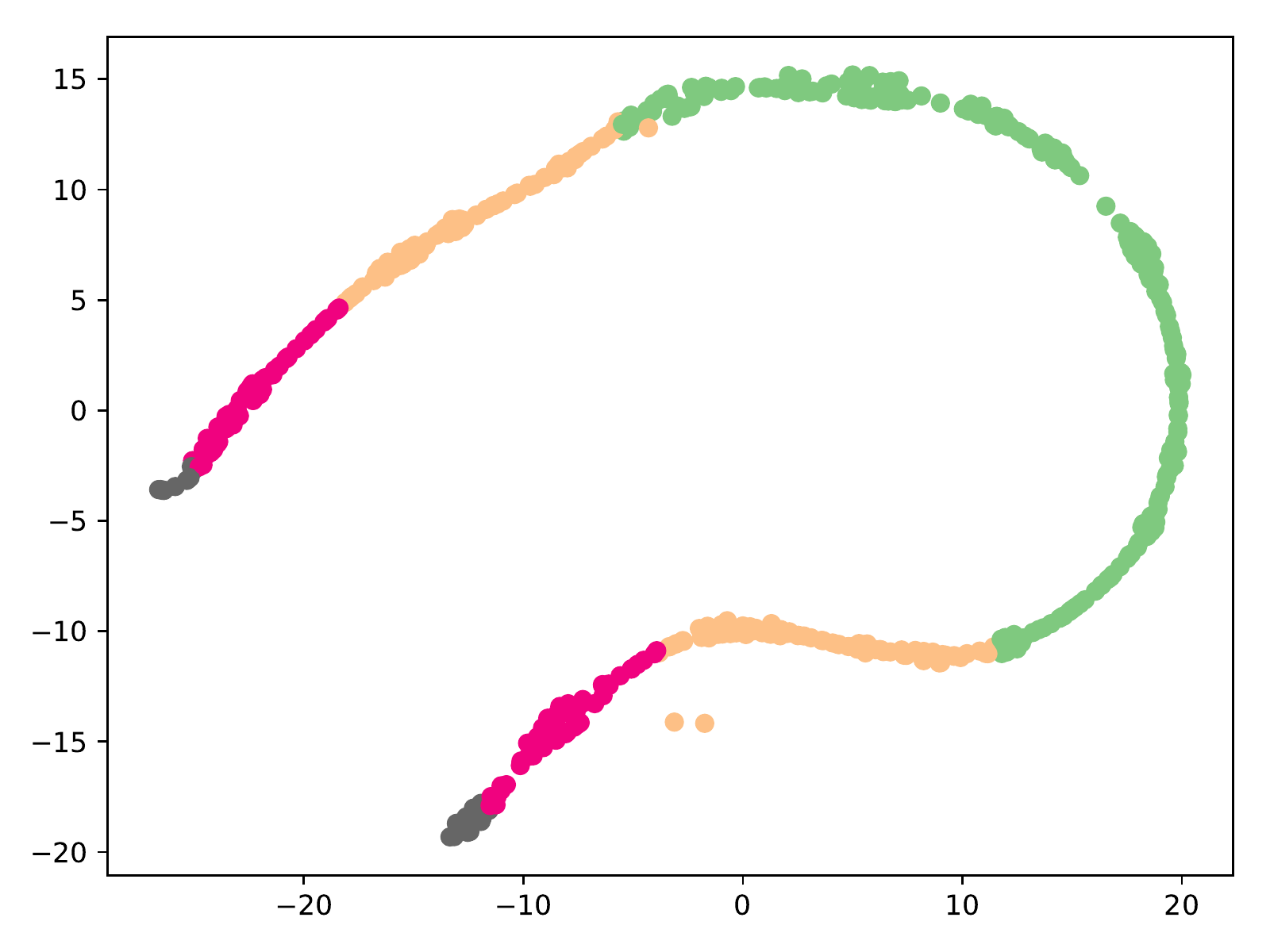}}
}

\centerline{
\subfloat{\includegraphics[width=1.15in,height=1.in]{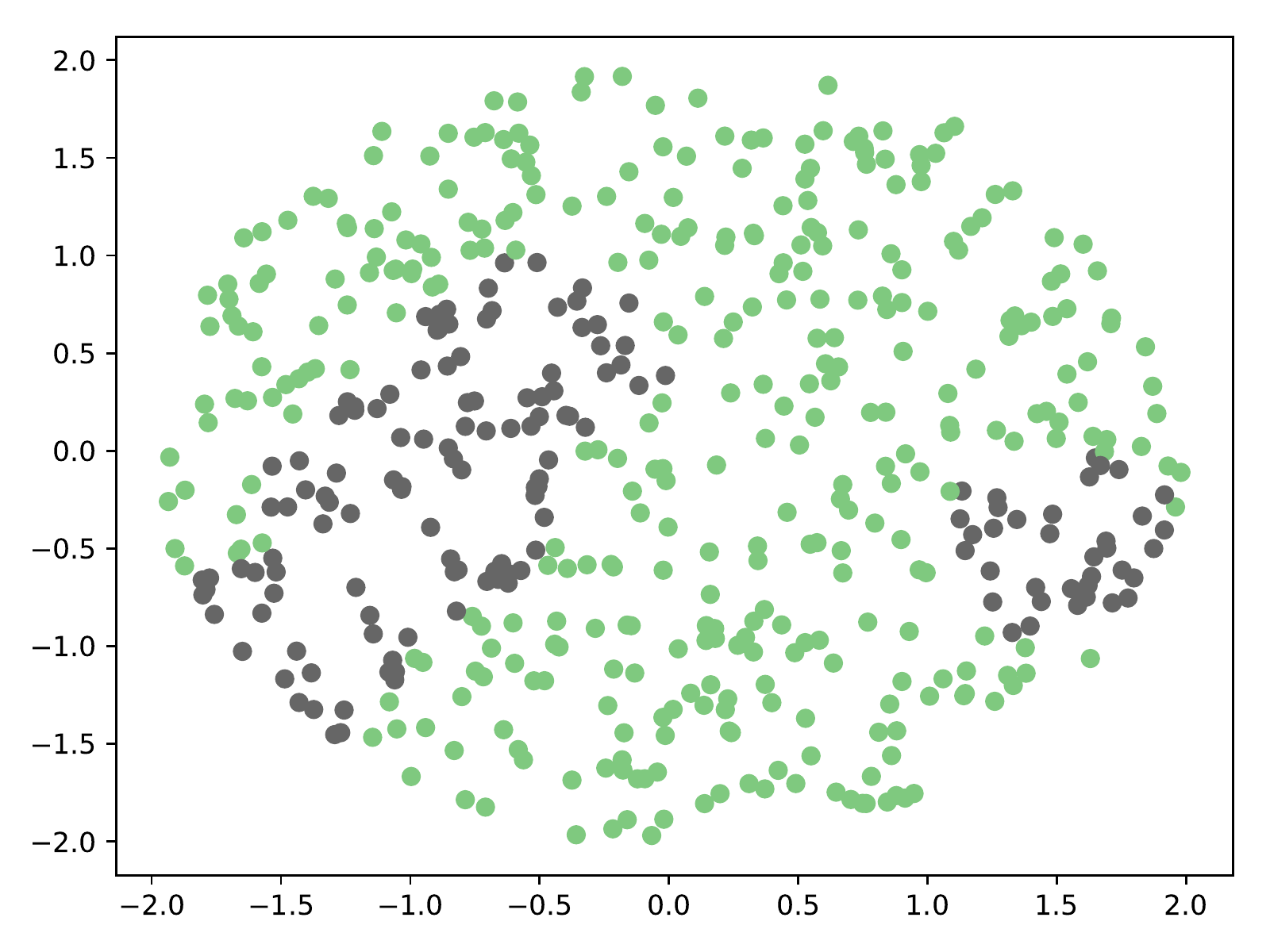}}
\subfloat{\includegraphics[width=1.15in,height=1.in]{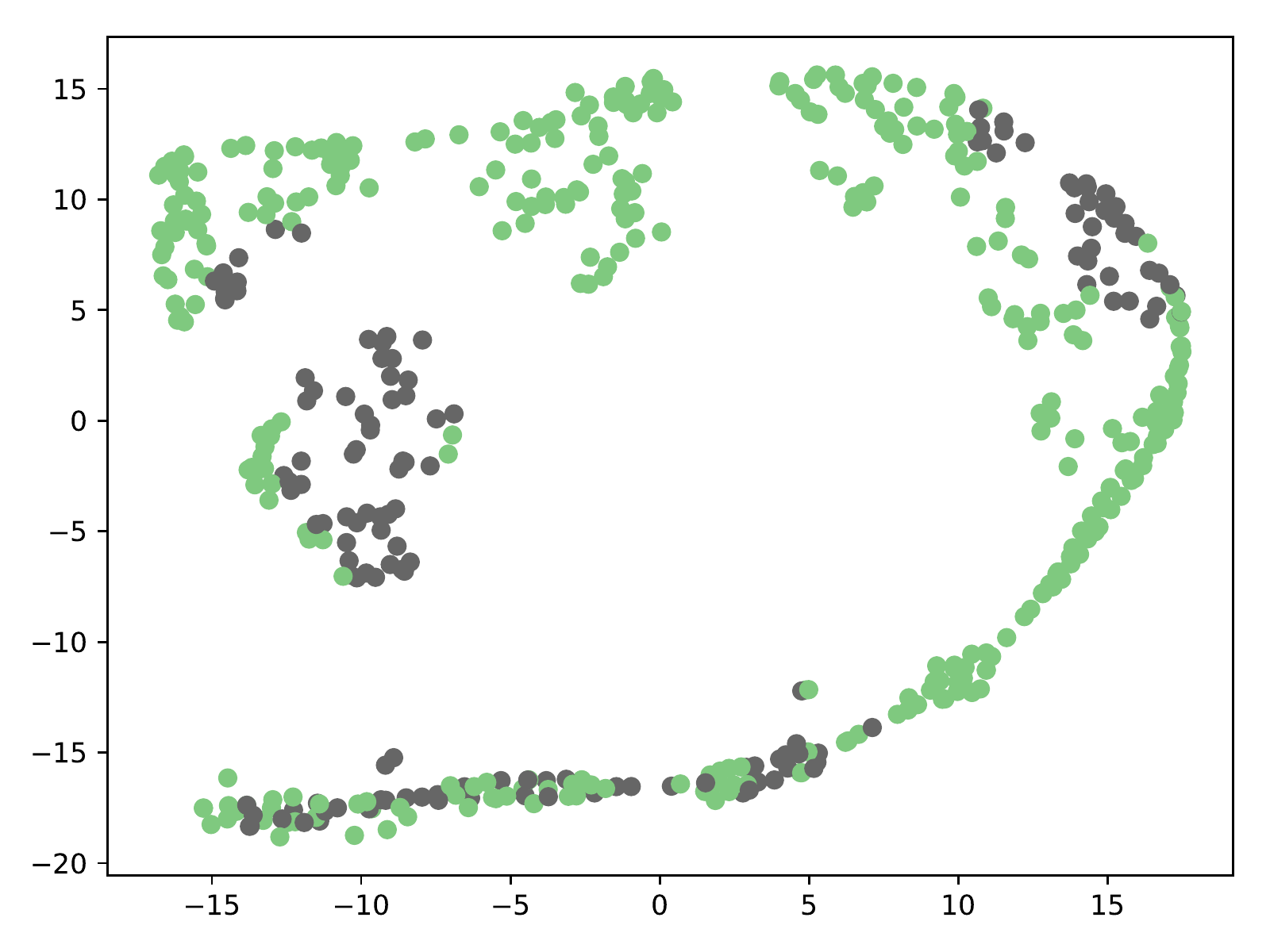}}
\subfloat{\includegraphics[width=1.15in,height=1.in]{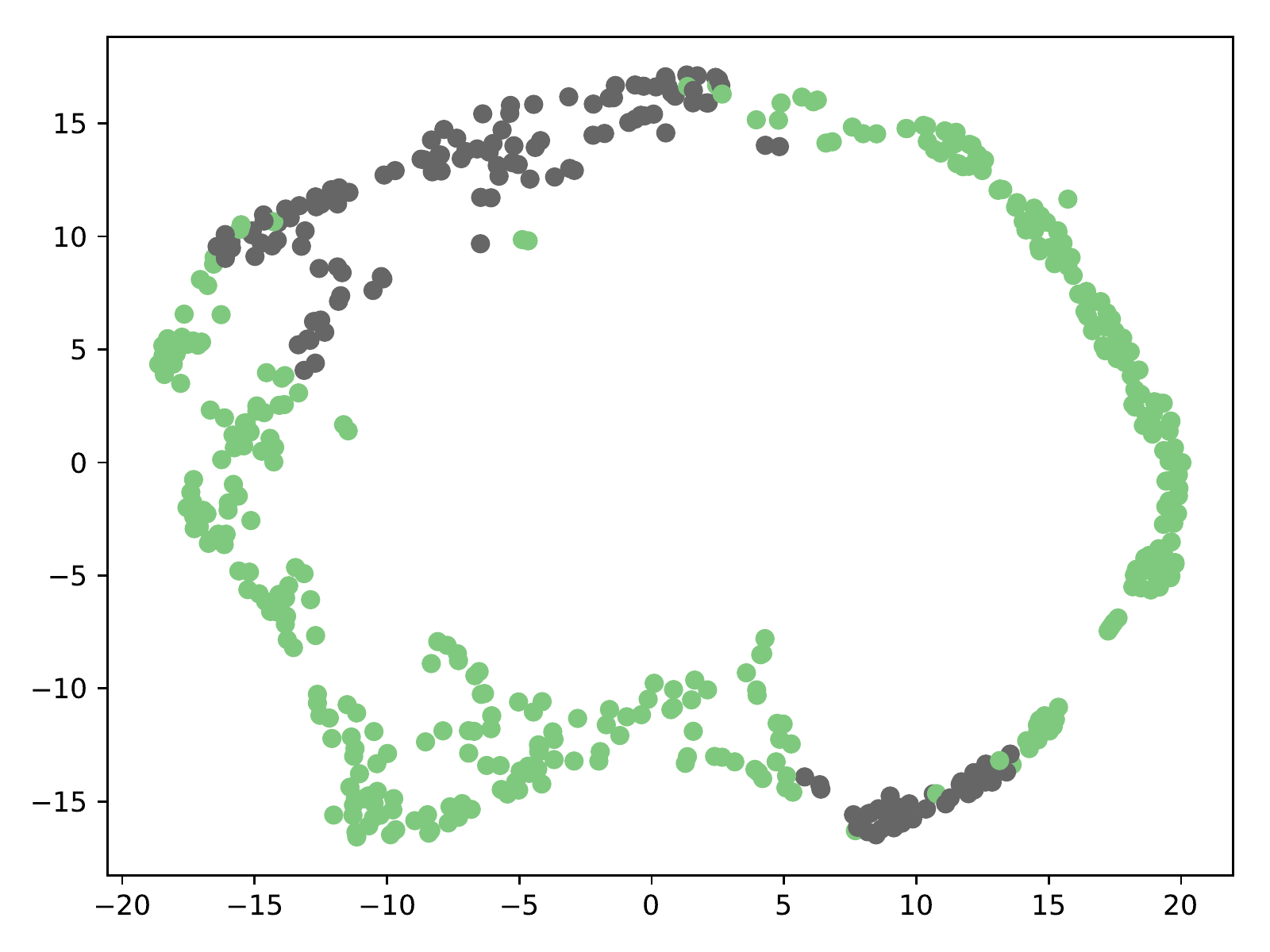}}
}

\centerline{
\subfloat{\includegraphics[width=1.15in,height=1.in]{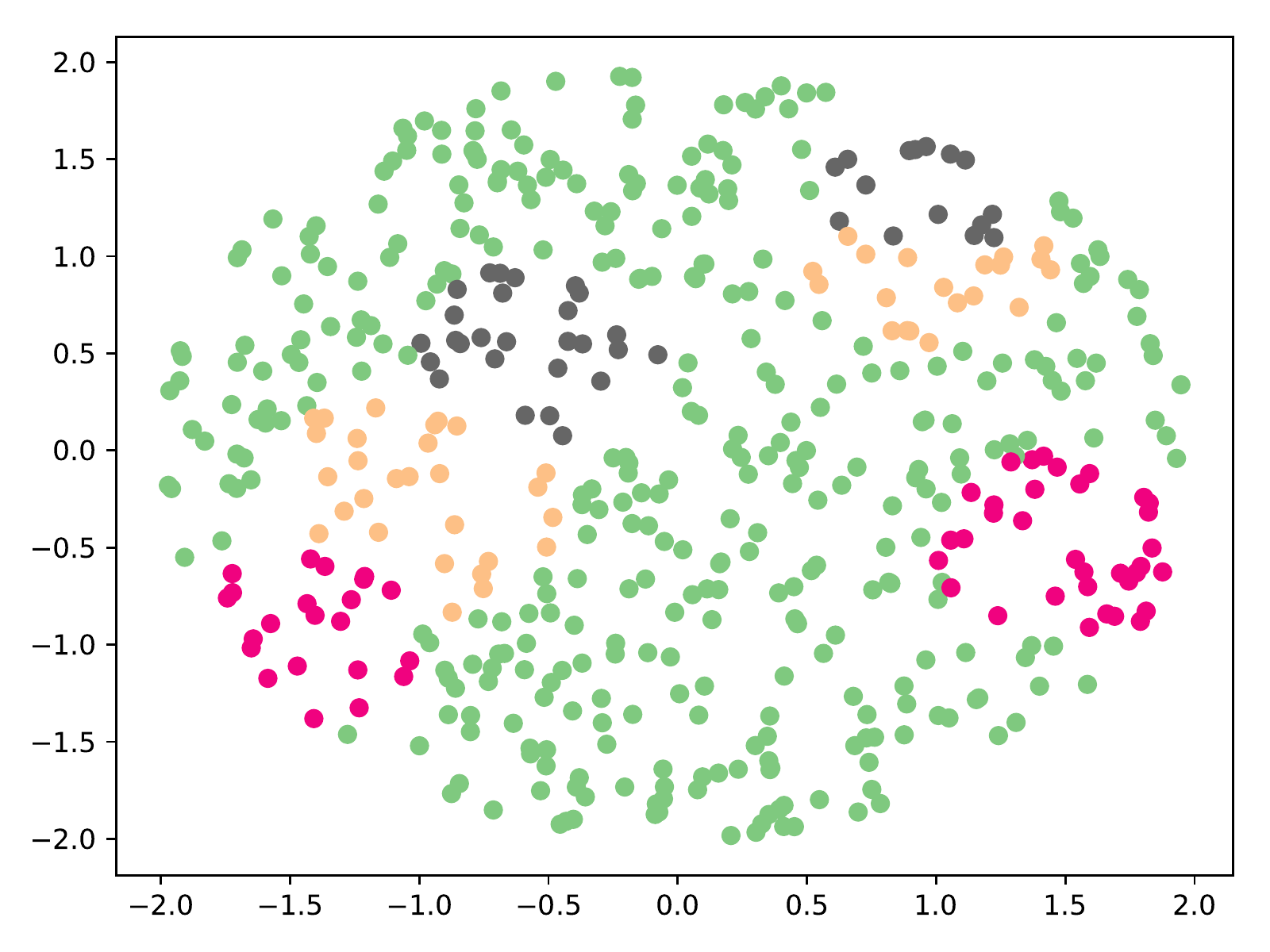}}
\subfloat{\includegraphics[width=1.15in,height=1.in]{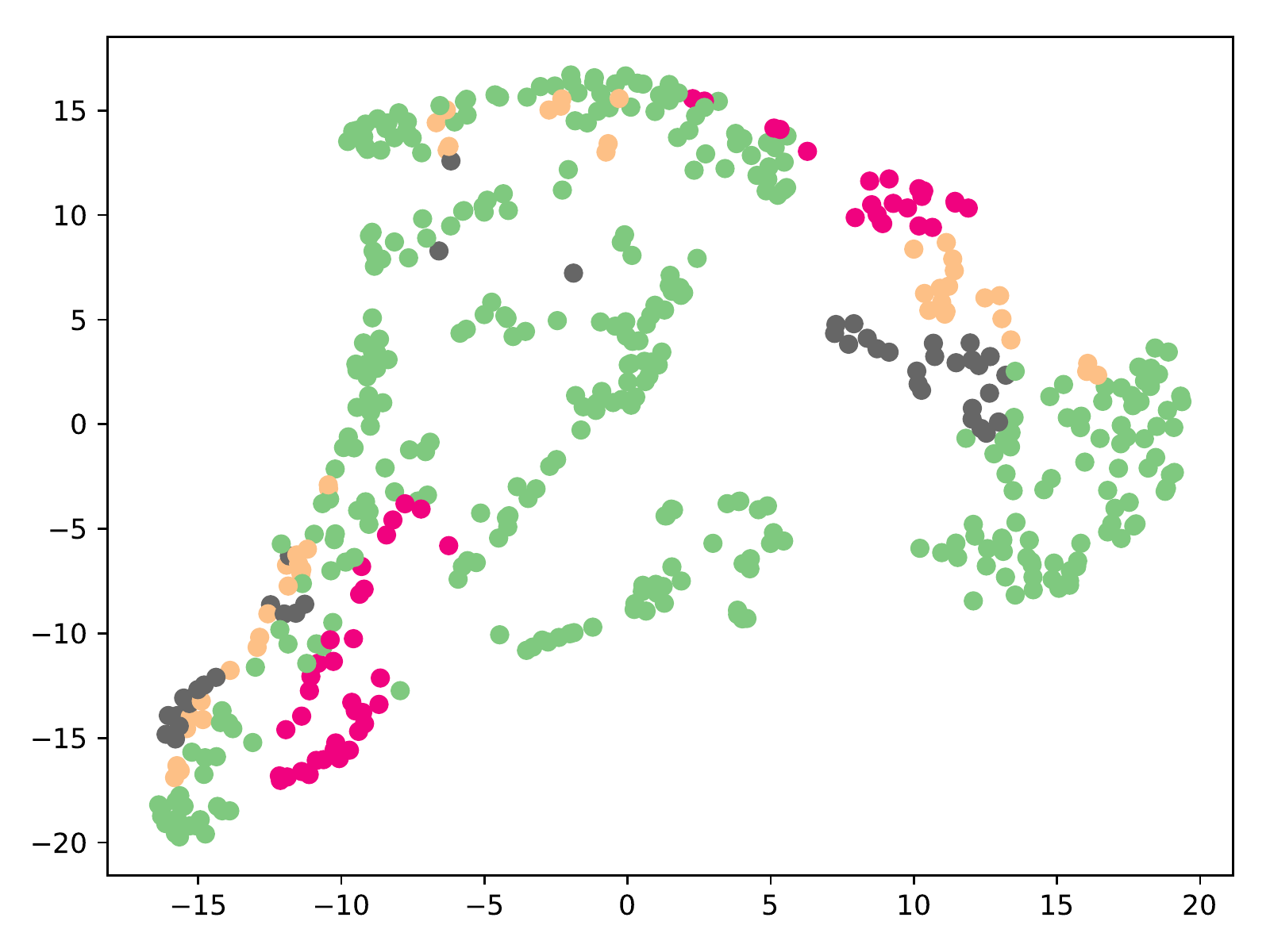}}
\subfloat{\includegraphics[width=1.15in,height=1.in]{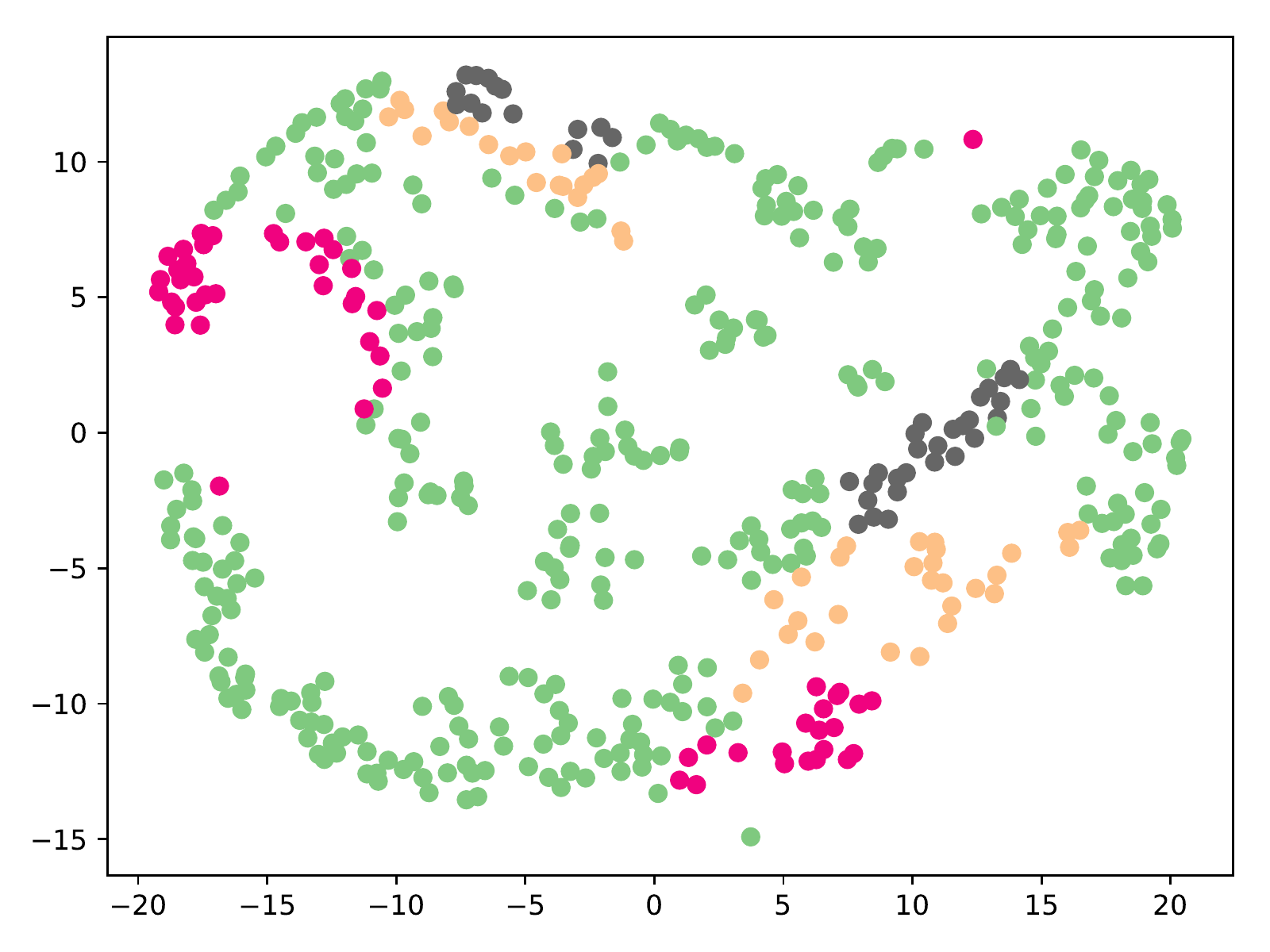}}
}

\caption{Visualization of experiments on additional synthetic binary and multi-class classification datasets. Figures on each row describe a single experiment. The dataset in each experiment is rendered on the left; the untrained $\mathcal{E}(\cdot)$ is in the middle; and the right column visualizes $\mathcal{E}(\cdot)$ after training. In the first two experiments, $\mathcal{E}: \mathbb{R}^2 \rightarrow \mathbb{R}^3$ is a neural network with 3 layers with $16$, $8$, and $3$ nodes each; in the remaining experiments, $\mathcal{E}: \mathbb{R}^2 \rightarrow \mathbb{R}^8$ contains just 2 layers with $16$ and $8$ nodes.}
\label{fig:synthetic_others}
\end{center}
\end{figure}

\begin{table*}[t]
\caption{Hyperparameters of the model for each dataset. We performed a grid search over the space of hyperparameters and identified the best combination by evaluation on the validation set.}
\label{table:hyperparameters}
\begin{center}
\begin{sc}
\begin{tabular}{llrcrrc}
\toprule
Dataset & Type of $\mathcal{F}$ & Max Trees & Max Tree Depth & $\sigma$ & Batch Size & Learning rate \\
\midrule
IMDB Movie Reviews & GBDT & 200 & 4 & .013 & 128 & 3e-4 \\
Yelp Reviews & Random Forest & 64 & 6 & .050 & 512 & 1e-4 \\
GLUE CoLA & GBDT & 64 & 5 & .100 & 64 & 3e-4 \\
GLUE STT2 & GBDT & 200 & 5 & .100 & 64 & 2e-4 \\
GLUE MRPC & Random Forest & 64 & 3 & .003 & 64 & 1e-3 \\
GLUE QQP & GBDT & 64 & 6 & .010 & 128 & 1e-4 \\
GLUE STSB & GBDT & 200 & 4 & .001 & 64 & 1e-4 \\
\bottomrule
\end{tabular}
\end{sc}
\end{center}
\end{table*}

The task would become trivial and the data could be modeled with a single decision tree (in fact, a single decision node), if the input to $\mathcal{F}$ was instead the difference $z = x_1 - x_2$: The split $z > 0$---axis-aligned in the embedding space, $\mathbb{R}$---would perfectly model the data. We therefore verify whether, through using our proposed method, the decision tree can direct an embedding function $\mathcal{E}: \mathbb{R}^2 \rightarrow \mathbb{R}$ to learn to project an input $(x_1, x_2)$ onto the real line using the transformation $\mathcal{E}(x_1, x_2) = x_1 - x_2$. For this experiment, the neural network serving as $\mathcal{E}$ is a single neuron (i.e., a linear transformation).

We have illustrated $\mathcal{E}$ for points in the $x_1$-$x_2$ plane in Figure~\ref{fig:synthetic_identity_line_decision_boundary}(b) and~\ref{fig:synthetic_identity_line_decision_boundary}(c) before and after training. From these figures and the weights of the neural network, we can confirm that $\mathcal{E}$ indeed takes the desired form.

\subsubsection{Other examples}
We have included results from more examples of synthetic datasets for binary and multi-class classification in Figure~\ref{fig:synthetic_others}. The figure illustrates the datasets in the left-most column, followed by the initial and trained embeddings. Where $\mathcal{E}$ projects to an embedding space with dimensionality larger than $3$, we use t-SNE~\cite{tsne} to render the data on the $2$-dimensional plane. From these figures, it is again evident that the model is able to learn embeddings that make it easier for the decision forest to separate the classes.

We set up the experiments above as follows. We generate $5000$ training points and $500$ test points for each dataset. To initialize $\mathcal{F}$, we set the max depth of each decision tree to 4 and generate 32 such trees randomly. $\mathcal{E}$ is a neural network consisting of either two layers with $16$-$8$ nodes each or three layers with $16$-$8$-$3$ nodes each, with the last number indicating the size of the output layer. To train the model end-to-end, we use a batch size of $512$. We use Adam~\cite{kingma:adam} to optimize the misclassification loss. We set the standard deviation in $\widetilde{F}_\sigma$ to 0.015. Finally, we use a subset of the training set as validation for early-stopping purposes.

\subsection{Benchmark Datasets}

We investigate RQ2 through experiments on a set of benchmark machine learning datasets. As noted earlier, we set the embedding function $\mathcal{E}$ to be the Universal Sentence Encoder~\cite{cer2018universal}. We then train a GBDT or a Random Forest over the output of $\mathcal{E}(\cdot)$. This initial state of our model, $\mathcal{F}\circ\mathcal{E}$ serves as a baseline. Our goal is then to fine-tune the embeddings and measure relative gains over the baseline.

The datasets we use in this section are as follows:
\begin{itemize}
    \item \textbf{IMDB Movie Reviews}~\cite{Maas2011IMDB}: The IMDB movie review dataset contains the raw text of $50,000$ movie reviews posted to IMDB, of which half are in the training set and the other half in the test set. The task is a sentiment classification of reviews into positive and negative classes.
    \item \textbf{Yelp Reviews}~\cite{zhang2015characterlevel}: Another binary sentiment classification dataset consisting of $560,000$ reviews for training and $38,000$ reviews for testing. The dataset was constructed by considering reviews with $1$ or $2$ stars as negative, and $3$ and $4$ as positive.
    \item \textbf{GLUE CoLA}~\cite{warstadt2018neural}: The Corpus of Linguistic Acceptability is a dataset of English sentences drawn from books and journal articles on linguistic theory and annotated with a binary label indicating whether each example is grammatical. There are $8,551$ training, $1,043$ validation, and $1,063$ test examples in this dataset.
    \item \textbf{GLUE STT2}~\cite{socher2013recursive}: The Stanford Sentiment Treebank consists of sentences from movie reviews and human annotations of their sentiment. The binary classification task is to predict sentence-level labels. The dataset contains $67,349$ training, $872$ validation, and $1,821$ test examples.
    \item \textbf{GLUE MRPC}~\cite{dolan2005automatically}: The Microsoft Research Paraphrase Corpus contains sentence pairs that are extracted from online news sources and are annotated by human judges with whether the sentences in each pair are semantically equivalent. The training set has $3,668$ examples, validation $408$, and test $1,725$.
    \item \textbf{GLUE QQP}~\cite{WinNT}: Similar to MRPC, the Quora Question Pairs dataset is a collection of question pairs from the community question-answering website Quora where the task is to determine whether a pair of questions are semantically equivalent. The training set consists of $363,849$ examples, validation of $40,430$, and test of $390,965$.
    \item \textbf{GLUE STSB}~\cite{cer2017semeval}: The Semantic Textual Similarity Benchmark consists of sentence pairs from news headlines, video and image captions, and natural language inference data. Each pair is annotated with a similarity score from $1$ to $5$. There are $5,749$ training, $1,500$ validation, and $1,379$ test examples in this dataset.
\end{itemize}

For each dataset, we perform a grid search over a set of possible values for model hyperparameters and choose the best combination based on an evaluation on the validation set. Table~\ref{table:hyperparameters} provides a summary to facilitate reproducibility.

Table~\ref{table:classification_results} reports the results of our experiments. We measure the accuracy of the model on the test dataset before and after fine-tuning. The accuracy before training represents the quality of $\mathcal{F}\circ\mathcal{E}$ where $\mathcal{E}$ is not tuned. The accuracy after training, on the other hand, is the quality of the model after fine-tuning of the embeddings. We also report the relative improvement between the initial state and the end state.

It is clear that on larger datasets (i.e., IMDB, Yelp, STT2 and QQP) fine-tuning the embeddings for the decision forest leads to significant improvements with an impressive 9.3\% increase in accuracy on the Yelp Review and GLUE QQP datasets. This trend does not hold on datasets with a very small number of training or validation examples, such as GLUE CoLA, MRPC, and STSB. In these instances, measurements on the validation set are bound to be less stable and more unreliable. Additionally, having very few examples in the training set makes the problem rather uninteresting, as the initial decision forest is likely to model the data more easily. Indeed in our experiments we observe that the model reaches 100\% accuracy on the training set very rapidly when there is insufficient data in the training and validation sets.

Setting aside the small datasets, the trend we observe here confirms that the decision forest benefits from fine-tuned embeddings. Simply using embeddings that are pre-trained for a different task in order to train a decision forest model leads to models with subpar quality. But fine-tuning the embeddings for the task at hand and at the direction of the decision forest leads to models with a higher quality.

It is worth noting that the accuracy reported in Table~\ref{table:classification_results} is unmistakably below the state of the art. But that is to be expected as we utilized a simple encoder for the embedding function $\mathcal{E}$. One may substitute the embedding layer with a more advanced deep neural network such as BERT~\cite{devlin2018bert}. But that is not the question we pursue here; instead we are interested only in studying whether a decision forest can direct and benefit from the fine-tuning of embeddings.

\begin{table}[t]
\caption{Accuracy on benchmark classification datasets measured on the test test. Initial accuracy is the accuracy of $\mathcal{F}\circ\mathcal{E}$ before the embeddings are fine-tuned. The ``fine-tuned'' column shows the accuracy after fine-tuning of embeddings has completed (i.e., when the model achieves the smallest loss on the validation set).}
\label{table:classification_results}
\begin{center}
\begin{sc}
\begin{tabular}{lccc}
\toprule
Dataset & Initial & Fine-tuned & $\Delta$ \\
\midrule
IMDB & 0.8436 & 0.8908 & +5.6\% \\
Yelp & 0.8717 & 0.9530 & +9.3\% \\
GLUE CoLA & 0.6894 & 0.6759 & -1.9\% \\
GLUE STT2 & 0.7764 & 0.8257 & +6.3\% \\
GLUE MRPC & 0.6887 & 0.6887 & 0 \\
GLUE QQP & 0.7598 & 0.8303 & +9.3\% \\
GLUE STSB & 0.2847 & 0.2887 & +1.4\% \\
\bottomrule
\end{tabular}
\end{sc}
\end{center}
\end{table}

%% file: conclusion.tex
\section{Discussion and Future Work}\label{sec:conclusion}

This work began by raising the following question: Can we learn or fine-tune embeddings directly to the advantage of a decision forest? We reviewed existing work in this domain but highlighted their shortcomings: Decision forests in existing models are either ancillary---lending support only in modeling tabular data---or are no longer axis-aligned. Furthermore, none of the existing methods result in decision forests that can simultaneously consume tabular and raw structured data (such as text and image). Finally, no prior work facilitates a fine-tuning of embeddings, a necessary task in many applications.

We set out to investigate this question and reported our solution in this work. Our proposal is straightforward and intuitive, and does much to address the drawbacks of prior work. In addition, it is flexible enough to allow the use of any existing Deep Learning technique for representation learning. It is also applicable to any decision forest learning algorithm.

Experiments on synthetic datasets demonstrated the feasibility and effectiveness of our proposed solution in learning embeddings that are meaningful to a decision forest and that make it easier for a decision forest to classify the transformed points. Further experiments on benchmark text classification datasets confirmed that fine-tuning pre-trained embeddings at the direction of a decision forest is not only possible with our framework, but also brings about significant improvements.

Now that we have introduced this general approach and presented its utility, we are interested in pursuing several other related directions in future. An important follow-up question, one that originally motivated us to pursue this work, is whether and in what ways our technique to learn embeddings for decision forests benefits state-of-the-art learning-to-rank algorithms.

A limitation of decision forest-based learning-to-rank algorithms such as LambdaMART~\cite{burges2005learning} or $\textsc{XE}_\textsc{NDCG}$~\cite{Bruch2019XEndcg} is that decision forests cannot consume raw text from queries or documents. Instead, these algorithms require extensive feature engineering. That, in turn, makes them inapplicable to or ineffective at datasets such as MS MARCO~\cite{bajaj2016msmarco}. This stands in contrast with Deep Learning-based methods that perform well on learning-to-rank datasets with raw text, but poorly on datasets with engineered features such as MSLR Web30K~\cite{DBLP:journals/corr/QinL13}. We hope to investigate whether our proposed method can help bridge this gap.

Another set of questions we hope to probe are on the topic of optimization and regularization. In this work, we used a Gaussian distribution to perturb the input to the decision forest. Notably, this perturbation affects all dimensions equally: The multivariate distribution had the identity matrix for covariance. But not all dimensions in the embedding space have the same input distribution. It may therefore be appropriate to use a noise distribution that is stretched in one dimension but squeezed in another. How the shape of the noise affects the optimization problem and the generalizability of the solution remains to be understood.

Finally, several aspects of our proposal would benefit from a closer examination. For example, the algorithm we used to generate random decision trees is rather straightforward; its effect on the final embeddings need to be studied further. Another example relates to the fact that a decision forest is effectively fixed after initialization (whether randomly initialized or trained). The question we hope to explore is what effect we will observe if we re-trained the decision forest periodically. We defer these investigations to future studies.